\documentclass{article}

\PassOptionsToPackage{numbers,sort}{natbib}

\usepackage[final]{neurips_2024}

\usepackage[utf8]{inputenc} 
\usepackage[T1]{fontenc}    
\usepackage{url}            
\usepackage{booktabs}       
\usepackage{amsfonts}       
\usepackage{nicefrac}       
\usepackage{microtype}      
\usepackage{xcolor}         
\usepackage{graphicx}

\usepackage{url}            
\usepackage{booktabs}       
\usepackage{amsfonts}       
\usepackage{nicefrac}       
\usepackage{microtype}      
\usepackage{xcolor}  
\usepackage{graphicx}
\usepackage{multirow}
\usepackage{color}
\usepackage{array}
\usepackage{algorithm}
\usepackage{algorithmic}
\usepackage{wrapfig}
\usepackage{amsmath}
\usepackage{makecell}
\usepackage{subfigure} 
\usepackage{colortbl}
\usepackage{xcolor}
\usepackage{caption}

\usepackage{amsmath,amsfonts,bm}

\usepackage{xspace}

\usepackage[backref]{hyperref} 
\hypersetup{
hidelinks,
colorlinks=true,
linkcolor=brown,
citecolor=brown
}

\title{Uncovering, Explaining, and Mitigating the Superficial Safety of Backdoor Defense}

\author{Rui Min\textsuperscript{1*}, \ Zeyu Qin\textsuperscript{1*}, \ Nevin L. Zhang\textsuperscript{1}, \ Li Shen, \ Minhao Cheng\textsuperscript{2}\\
\textsuperscript{1}Hong Kong University of Science and Technology,
\textsuperscript{2}Pennsylvania State University\\
\texttt{\{rminaa, zeyu.qin\}@connect.ust.hk}, \texttt{lzhang@cse.ust.hk}\\
\texttt{mathshenli@gmail.com},
\texttt{minhaocheng@ust.hk}
}

\begin{document}

\maketitle

\begin{abstract}

   Backdoor attacks pose a significant threat to Deep Neural Networks (DNNs) as they allow attackers to manipulate model predictions with backdoor triggers. To address these security vulnerabilities, various backdoor purification methods have been proposed to purify compromised models. Typically, these purified models exhibit low Attack Success Rates (ASR), rendering them resistant to backdoored inputs. However, \textit{Does achieving a low ASR through current safety purification methods truly eliminate learned backdoor features from the pretraining phase?}
   In this paper, we provide an affirmative answer to this question by thoroughly investigating the \textit{Post-Purification Robustness} of current backdoor purification methods. We find that current safety purification methods are vulnerable to the rapid re-learning of backdoor behavior, even when further fine-tuning of purified models is performed using a very small number of poisoned samples. Based on this, we further propose the practical Query-based Reactivation Attack (QRA) which could effectively reactivate the backdoor by merely querying purified models.
   We find the failure to achieve satisfactory post-purification robustness stems from the insufficient deviation of purified models from the backdoored model along the backdoor-connected path. To improve the post-purification robustness, we propose a straightforward tuning defense, Path-Aware Minimization (PAM), which promotes deviation along backdoor-connected paths with extra model updates. Extensive experiments demonstrate that PAM significantly improves post-purification robustness while maintaining a good clean accuracy and low ASR. Our work provides a new perspective on understanding the effectiveness of backdoor safety tuning and highlights the importance of faithfully assessing the model's safety.

\end{abstract}

\renewcommand{\thefootnote}{\fnsymbol{footnote}}
\footnotetext[1]{Equal contribution. Correspondence to: Zeyu Qin (\texttt{zeyu.qin@connect.ust.hk}), Li Shen, Minhao Cheng.}
\renewcommand{\thefootnote}{\arabic{footnote}}

\section{Introduction}
\label{sec:intro}

Backdoor attacks~\cite{gu2019badnets, chen2017targeted, bagdasaryan2020backdoor} have emerged as one of the most significant  concerns~\cite{kumar2020adversarial,carlini2022poisoning,carlini2023poisoning,hubinger2024sleeper} in deep learning. These attacks involve the insertion of malicious backdoor triggers into the training set, which can be further exploited to manipulate the behavior of the model during the inference stage. To defend against these threats, researchers have proposed various safety tuning methods~\cite{wu2021adversarial, li2023reconstructive, min2024towards, zhu2023enhancing, wang2019neural, xu2023towards, hubinger2024sleeper} to purify well-trained backdoored models. These methods can be easily incorporated into the existing model deployment pipeline and have demonstrated state-of-the-art effectiveness in reducing the Attack Success Rate (ASR) of backdoored models~\cite{wu2022backdoorbench, min2024towards}.

However, a critical question arises: \textit{does achieving a low Attack Success Rate (ASR) through current safety tuning methods genuinely indicate the complete removal of learned backdoor features from the pretraining phase?} If the answer is no, this means that the adversary may still easily reactivate the implanted backdoor from the residual backdoor features lurking within the purified model, thereby exerting insidious control over the model's behavior.  This represents a significant and previously unacknowledged safety concern, suggesting that current defense methods may only offer \textit{superficial safety} \cite{anwar2024foundational}. Moreover, if an adversary can successfully re-trigger the backdoor, it raises another troubling question: how can we assess the model's robustness against such threats? This situation underscores the urgent need for a more comprehensive and faithful evaluation of the model's safety.

In this work, we provide an affirmative answer to these questions by thoroughly investigating the \textbf{Post-Purification Robustness} of state-of-the-art backdoor safety tuning methods. Specifically, we employ the \textit{Retuning Attack} (RA) \cite{qi2023revisiting,tarun2023fast} where we first retune the purified models using an extremely small number of backdoored samples and tuning epochs. Our observations reveal that current safety purification defense methods quickly reacquire backdoor behavior after just a few epochs, resulting in significantly high ASR levels. In contrast, the clean model (which does not have backdoor triggers inserted during the pretraining phase) and \textit{Exact Purification} (EP)—which fine-tunes models using real backdoored samples with correct labels during safety purification, maintain a low ASR even after the RA. This discrepancy suggests that existing safety tuning methods do not thoroughly eliminate the learned backdoor,  creating a \textit{superficial impression of backdoor safety}.
Since the vulnerability revealed by the Retuning Attack (RA) relies on the use of retuned models, we further propose the more practical Query-based Reactivation Attack (QRA). This attack is capable of generating sample-specific perturbations that can trigger the backdoor in purified models, which were previously believed to have eliminated such threats, simply by querying these purified models.

To understand the inherent vulnerability of current safety purification methods concerning post-purification robustness, we further investigate the factors contributing to the disparity in post-purification robustness between EP and other methods. To this end, we utilize Linear Mode Connectivity (LMC) \cite{neyshabur2020being,frankle2020linear} as a framework for analysis. We find that \textit{EP not only produces a solution with low ASR like other purification methods but also pushes the purified model further away from the backdoored model along the backdoor-connected path, resulting in a more distantly robust solution.} As a result, it becomes challenging for the retuning attack to revert the EP model back to the basin with high ASR where the compromised model is located. Inspired by our findings, we propose a simple tuning defense method called Path-Aware Minimization (PAM) to enhance post-purification robustness. By using reversed backdoored samples as a proxy to measure the backdoored-connected path, PAM updates the purified model by applying gradients from a model interpolated between the purified and backdoored models. This approach helps identify a robust solution that further deviates our purified model from the backdoored model along the backdoor-connected path. Extensive experiments have demonstrated that PAM achieves improved post-purification robustness, retaining a low ASR after RA across various settings. To summarize, our contributions are:
\begin{itemize}
    \item Our work first offers a new perspective on understanding the effectiveness of current backdoor safety tuning methods. Instead of merely focusing on the commonly used Attack Success Rate, we investigate the Post-Purification Robustness of the purified model to enhance our comprehensive understanding of backdoor safety in deep learning models. 
    
    \item We employ the Retuning Attack by retuning purified models on backdoored samples to assess the post-purification robustness. Our primary observations reveal that current safety purification methods are vulnerable to RA, as evidenced by a rapid increase in the ASR. Furthermore, we propose the more practical Query-based Reactivation Attack, which can reactivate the implanted backdoor of purified models solely through model querying.
    
    \item We analyze the inherent vulnerability of current safety purification methods to the RA through Linear Mode Connectivity and attribute the reason to the insufficient deviation of purified models from the backdoored model along the backdoor-connected path. Based on our analysis, we propose Path-Aware Minimization, a straightforward tuning-based defense mechanism that promotes deviation by performing extra model updates using interpolated models along the path. Extensive experiments verify the effectiveness of the PAM method.

\end{itemize}

\section{Related Work}

\paragraph{Backdoor Attacks.} Backdoor attacks aim to manipulate the backdoored model to predict the target label on samples containing a specific backdoor trigger while behaving normally on benign samples. They can be roughly divided into two categories \cite{wu2022backdoorbench}: \textbf{(1)} Data-poisoning attacks: the attacker inserts a backdoor trigger into the model by manipulating the training sample $(\bm{x}, \bm{y}) \in (\mathcal{X}, \mathcal{Y})$, such as adding a small patch to a clean image $\bm{x}$ and change the sample's label to an attacker-designated target label $\bm{y}_t$ \cite{goldblum2022dataset,gu2019badnets,li2021invisible,chen2017targeted,carlini2022poisoning,turner2019label}; \textbf{(2)} Training-control attacks: the attacker has control over both the training process and the training data simultaneously~\cite{nguyen2021wanet}. Note that data-poisoning attacks are more practical in real-world scenarios as they make fewer assumptions about the attacker's capabilities~\cite{carlini2023poisoning,shejwalkar2022back,goldblum2022dataset} and have resulted in increasingly serious security risks~\cite{qin2023revisiting,carlini2022poisoning}. In this work, \textit{we focus on data-poisoning backdoor attacks}. 

\paragraph{Backdoor Defense.} Existing backdoor defense strategies could be broadly categorized into robust pretraining~\cite{huang2022backdoor,li2021anti} and robust fine-tuning methods~\cite{wu2021adversarial,zeng2022adversarial,min2024towards,zhu2023enhancing}. Robust pretraining aims to prevent the learning of backdoor triggers during the pretraining phase. However, these methods often suffer from accuracy degradation and can significantly increase model training costs, making them impractical for large-scale applications. In contrast, robust purification methods focus on removing potential backdoor features from a well-trained model. Generally, purification techniques involve reversing potential backdoor triggers \cite{wang2019neural,wang2022universal,wang2022unicorn,xu2023towards} and applying fine-tuning or pruning to address backdoors using a limited amount of clean data \cite{liu2018fine,wu2021adversarial,zhu2023enhancing,min2024towards}. While these purification methods reduce training costs, they also achieve state-of-the-art defense performance \cite{zhu2023enhancing,min2024towards}. Therefore, in this work, \textit{we mainly focus on evaluations of robust purification methods against backdoor attacks.}

\paragraph{Loss Landscape and Linear Mode Connectivity.}  Early works \cite{draxler2018essentially,garipov2018loss,izmailov2018averaging} conjectured and empirically verified that different DNN loss minima can be connected by low-loss curves.
In the context of the pretrain-fine-tune paradigm, \citet{neyshabur2020being} observe that the pretrained weights guide purified models to the same flat basin of the loss landscape, which is close to the pretrained checkpoint. \citet{frankle2020linear} also observe and define the linear case of mode connectivity, \textit{Linear Mode Connectivity} (LMC). LMC refers to the absence of the loss barrier when interpolating linearly between solutions that are trained from the same initialization. The shared initialization can either be a checkpoint in early training \cite{frankle2020linear} or a pretrained model \cite{neyshabur2020being}. \textit{Our work is inspired by \cite{neyshabur2020being} and also utilizes LMC to investigate the properties of purified models in relation to backdoor safety.}

\paragraph{Deceptive AI and Superficial Safety.} Nowadays, DNNs are typically pretrained on large-scale datasets, such as web-scraped data, primarily using next-token prediction loss \cite{tomfewshot}, along with simple contrastive \cite{chen2020simple} and classification \cite{alexnet2012} objective. While these simplified pretraining objectives can lead to the learning of rich and useful representations, they may also result in deceptive behaviors that can mislead humans \cite{elk_report}. One such deceptive behavior is the presence of backdoors \cite{mad_report,hubinger2024sleeper}. A compromised model can be indistinguishable from a normal model to human supervisors, as both behave similarly in the absence of the backdoor trigger. To address this critical safety risk, researchers propose post-training alignment procedures, such as safety fine-tuning \cite{ouyang2022training,wu2022backdoorbench}. However, several studies indicate that the changes from fine-tuning are \textit{superficial} \cite{zhou2024lima,lubana2023mechanistic}. As a result, these models retain harmful capabilities and knowledge from pretraining, which can be elicited by harmful fine-tuning \cite{qi2023fine,yang2023shadow} or specific out-of-distribution (OOD) inputs \cite{wei2024jailbroken,yuan2024gpt}. \textit{We study this phenomenon in the context of backdoor threats and offer a deeper understanding along with mitigation strategies.}

\section{Revealing Superficial Safety of Backdoor Defenses by Accessing Post-purification Robustness}
\label{sec:revisiting-evaluations}

While current backdoor purification methods can achieve a very low Attack Success Rate (ASR) against backdoor attacks, this does not necessarily equate to the complete elimination of inserted backdoor features. Adversaries may further exploit these residual backdoor features to reconstruct and reactivate the implanted backdoor, as discussed in Section~\ref{sec:black-box-evaluation}.
This is particularly important because purified models are often used in various downstream scenarios, such as customized fine-tuning \cite{qi2023fine} for critical tasks \cite{hubinger2024sleeper}. Therefore, it is crucial to provide a way to measure the robustness of purified models in defending against backdoor re-triggering, which we define as \textbf{"Post-Purification Robustness"}.

In this section, we first introduce a simple and straightforward strategy called the Retuning Attack (RA) to assess post-purification robustness. Building on the RA, we then present a practical threat known as the Query-based Reactivation Attack (QRA), which exploits the vulnerabilities in post-purification robustness to reactivate the implanted backdoor in purified models, using only model querying. First, we will introduce the preliminaries and evaluation setup.

\subsection{Problem Setup}
\label{sec:problem-setup}

\paragraph{Backdoor Purification.}
In this work, we focus on the poisoning-based attack due to its practicality and stealthiness. We denote the original training dataset as $\mathcal{D}_{T} \subset (\mathcal{X}, \mathcal{Y})$. A few training examples $(\bm{x}, \bm{y}) \in \mathcal{D}_{T}$ have been transformed by attackers into poisoned examples $(\bm{x}_{p}, \bm{y}_t)$, where $\bm{x}_{p}$ is poisoned example with inserted trigger and a target label $\bm{y}_t$. Following previous works \cite{liu2018fine,wu2021adversarial,wu2022backdoorbench,qin2023revisiting,min2024towards}, only a limited amount of clean data $D_{t}$ are used for fine-tuning or pruning. For trigger-inversion methods \cite{wang2019neural,xu2023towards}, we denote the reversed backdoored samples obtained through reversing methods as $(\bm{x}_r, \bm{y})\in\mathcal{D}_{r}$. We evaluate several mainstreamed purification methods, including pruning-based defense \textit{ANP}~\cite{wu2021adversarial}; robust fine-tuning defense \textit{I-BAU}~\cite{zeng2022adversarial} (referred to as BAU for short), \textit{FT-SAM}~\cite{zhu2023enhancing} (referred to as SAM for short), \textit{FST}~\cite{min2024towards}, as well as the state-of-the-art trigger-reversing defense \textit{BTI-DBF}~\cite{xu2023towards} (referred to as BTI for short). BTI purifies the backdoored model by using both reversed backdoored samples $\mathcal{D}_{r}$ and the clean dataset $\mathcal{D}_{t}$ while the others use solely the clean dataset $\mathcal{D}_{t}$. 
We also include \textit{exact purification (EP)} that assumes that the defender has full knowledge of the exact trigger and fine-tunes the models using real backdoored samples with correct labels $(\bm{x}_{p}, \bm{y})$.

\paragraph{Attack Settings.} Following \cite{min2024towards}, we evaluate four representative data-poisoning backdoors including three dirty-label attacks (BadNet~\cite{gu2019badnets}, Blended~\cite{chen2017targeted}, SSBA~\cite{li2021invisible}), and one clean-label attack (LC~\cite{turner2019label}). 
All experiments are conducted on BackdoorBench~\cite{wu2022backdoorbench}, a widely used benchmark for backdoor learning. 
We employ three poisoning rates, $10\%, 5\%$, and $1\%$ (in \textit{Appendix}) for backdoor injection and conduct experiments on three widely used image classification datasets, including CIFAR-10~\cite{krizhevsky2009learning}, Tiny-ImageNet~\citep{chrabaszcz2017downsampled}, and CIFAR-100~\cite{krizhevsky2009learning}. For model architectures, we following~\cite{min2024towards}, and adopt the ResNet-18, ResNet-50~\cite{He_2016_CVPR}, and DenseNet-161~\cite{huang2017densely} on CIFAR-10. For CIFAR-100 and Tiny-ImageNet, we adopt pretrained ResNet-18 on \textit{ImageNet1K} to obtain high clean accuracy as suggested by~\cite{xu2023towards,min2024towards}. More details about experimental settings are shown in \textit{Appendix}~\ref{app:exp_settings}.

\paragraph{Evaluation Metrics.} Following previous backdoor works, we take two evaluation metrics, including \textit{Clean Accuracy (C-Acc)} (i.e., the prediction accuracy of clean samples) and \textit{Attack Success Rate (ASR)} (i.e., the prediction accuracy of poisoned samples to the target class) where a lower ASR indicates a better defense performance. We further adopt \textit{O-ASR} and \textit{P-ASR} metrics. The O-ASR metric represents the defense performance of original defense methods,
while the P-ASR metric indicates the ASR after applying the RA or QRA.

\subsection{Purified Models Are Vulnerable to Retuning Attack}
\label{sec:retuning-attack}

Our objective is to investigate whether purified models with low ASR completely eliminate the inserted backdoor features. To accomplish this, it is essential to develop a method for assessing the degree to which purified models have indeed forgotten these triggers. In this section, \textit{we begin with a white-box investigation where the attacker or evaluator has access to the purified model's parameters.} Here we introduce a simple tuning-based strategy named the \textbf{Retuning Attack (RA)} \cite{qi2023fine,tarun2023fast} to conduct an initial evaluation. Specifically, we construct a dataset for model retuning, which comprises a few backdoored samples (less than $1\%$ of backdoored samples used during the training process). To maintain C-Acc, we also include benign samples from the training set, resulting in a total RA dataset with $1000$ samples. We subsequently retune the purified models using this constructed dataset through a few epochs ($5$ epochs in our implementation). This approach is adopted because a clean model can not be able to learn a backdoor; thus, if the purified models quickly regain ASR during the retuning process, it indicates that some residual backdoor features still exist in these purified models. Implementation details of the RA can be found in the \textit{Appendix}~\ref{app:implement_details}.

\begin{figure}
\centerline{\includegraphics[width=1.05\textwidth]{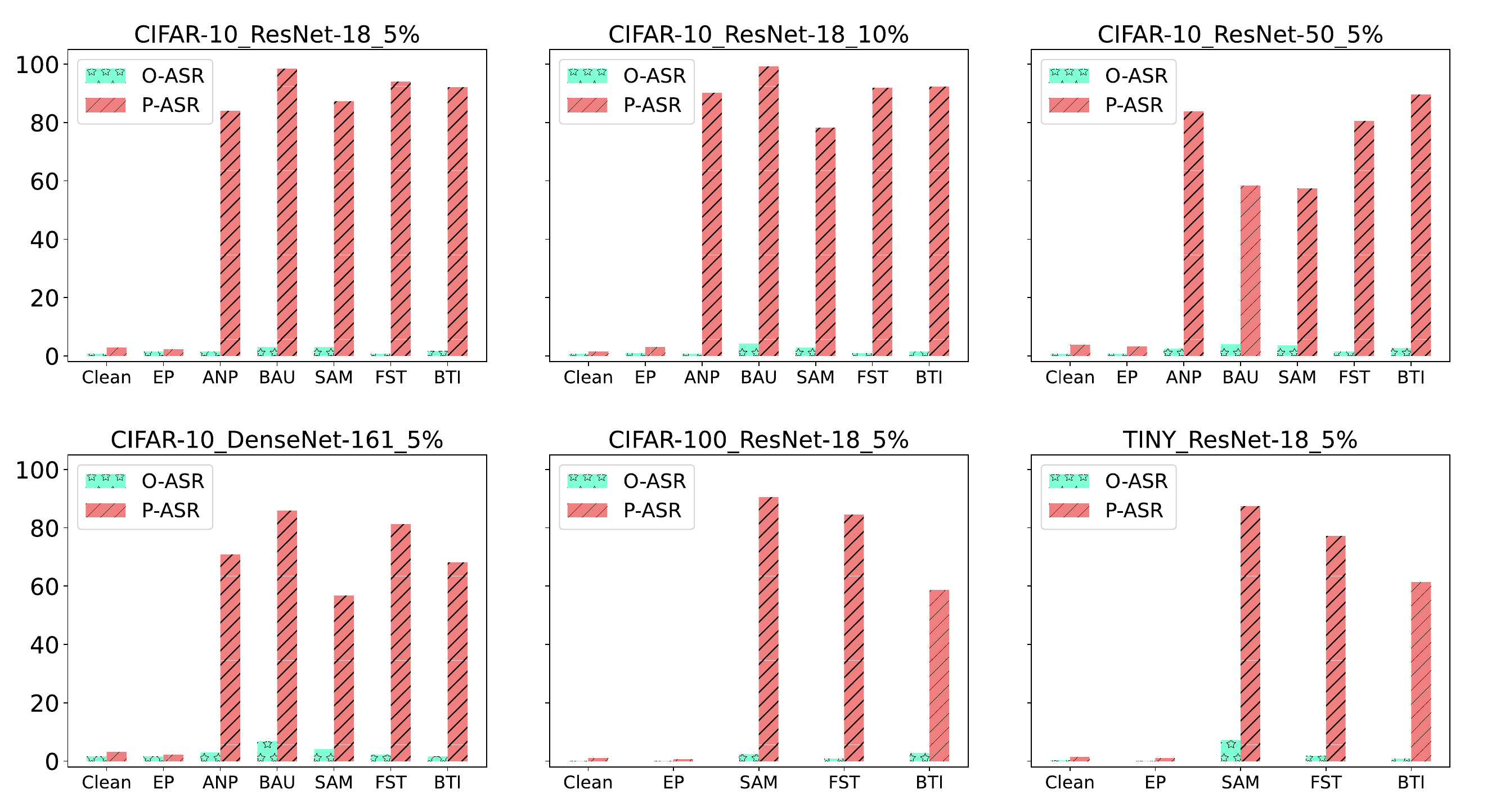}}
    \caption{\small The robustness performance against various attack settings. The title consists of the used dataset, model, and poisoning rate. The \textit{O-ASR} metric represents the defense performance of original defense methods, while the \textit{P-ASR} metric indicates the ASR after applying the RA. All metrics are measured in percentage (\%) Here we report the average results among backdoor attacks and defer more details in~\textit{Appendix}~\ref{app:detailed_ra}.}
    \label{figure:revisiting-results}
\end{figure}

As shown in Figure~\ref{figure:revisiting-results}, we observe that \textit{despite achieving very low ASR, all purification methods quickly recover backdoor ASR with Retuning Attack.} Their quickly regained ASR presents a stark contrast to that of clean models and remains consistent across different datasets, model architectures, and poisoning rates. Note that the pruning method (ANP) and the fine-tuning method (FST), which achieve state-of-the-art defense performance, still exhibit vulnerability to RA, with an average recovery of approximately $82\%$ and $85\%$ ASR, respectively. In stark contrast, the EP method stands out as \textit{it consistently maintains a low ASR even after applying RA, demonstrating exceptional post-purification robustness.} Although impractical with full knowledge of the backdoor triggers, the EP method validates the possibility of maintaining a low attack success rate to ensure post-purification robustness against RA attacks.

Moreover, this evident contrast highlights the significant security risks associated with current backdoor safety tuning methods. While these methods may initially appear robust due to a significantly reduced ASR, they are fundamentally vulnerable to backdoor reactivation, which can occur with just a few epochs of model tuning. This superficial safety underscores the urgent need for more comprehensive evaluations to ensure lasting protection against backdoor attacks. It is crucial to implement faithful evaluations that thoroughly assess the resilience of purified models, rather than relying solely on superficial metrics, to truly safeguard against the persistent threat of backdoor vulnerabilities.

\subsection{Reactivating Backdoor on Purified Models through Queries} 

\label{sec:black-box-evaluation}

Although our previous experiments on RA demonstrate that current purification methods insufficiently eliminate learned backdoor features, it is important to note that the success of this tuning-based method relies on the attackers' capability to change purified models' weights. This is not practical in a real-world threat model. To address this limitation, we propose \textbf{Query-based Reactivation Attack (QRA)}, which generates sample-specific perturbations that can reactivate the backdoor using only model querying. Specifically, instead of directly retuning purified models, QRA captures the parameter changes induced by the RA process and translates them into input space as perturbations. These perturbations can then be incorporated into backdoored examples, facilitating the successful reactivation of backdoor behaviors in purified models.

\begin{figure}

\centerline{\includegraphics[width=1.05\textwidth]{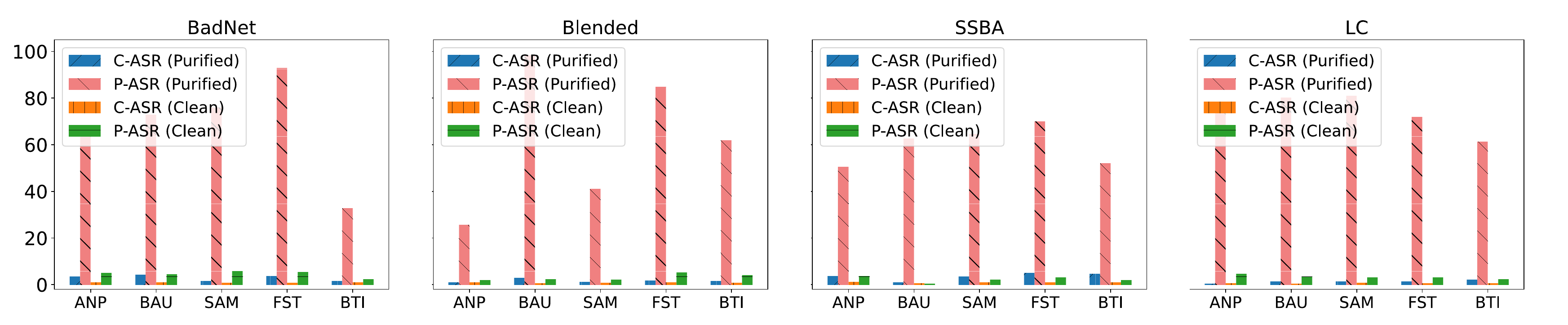}}
    \caption{\small Experimental results of QRA on both the \textit{purified} and \textit{clean} models against four types of backdoor attacks. We evaluate the QRA on CIFAR-10 with ResNet-18 and the poisoning rate is set to $5\%$. Additional results of QRA are demonstrated in \textit{Appendix}~\ref{app:detailed_qra}.}
    \label{fig:ba}
\end{figure}

To effectively translate the parameter changes into the input space, it is crucial to ensure that when applying the perturbation generated by QRA, the output of the purified model on perturbed inputs should be aligned with that of the post-RA model on original inputs without perturbations. Formally, we denote the purified model as $\bm{f}(\bm{W}_p;\bm{x})$ , the model after RA as $\bm{f}(\bm{W}_{ra};\bm{x})$ and their corresponding logit output as $\bm{l}(\bm{W}_p;\bm{x})$ and $\bm{l}(\bm{W}_{ra};\bm{x})$. Our QRA aims to learn a perturbation generator $\bm{\phi}(\bm{\theta}; \bm{x}): R^d \rightarrow [-1,1] ^d $,  to produce perturbation $\bm{\phi}(\bm{\theta}; \bm{x})$ for each input $\bm{x}$. We formulate this process into the following optimization problem:
\begin{equation}
\label{eq1}
    \min_{\bm{\theta}} \Big\{\mathbb{E}_{(\bm{x})\sim\mathcal{D}_c}[\mathcal{S}(\bm{l}(\bm{W}_{ra};\bm{x}), \bm{l}(\bm{W}_p;\epsilon*\bm \phi(\bm{\theta}; \bm{x})+\bm{x}))] \Big\}.
\end{equation}
Here $\mathcal{S}$ is the distance metric between two output logits, $\mathcal{D}_c$ is a compositional dataset incorporating both benign and backdoored samples, and $\epsilon$ controls the strength of perturbation ($\epsilon=16/255$ in our implementation). We utilize the Kullback–Leibler (KL) divergence \cite{cover1999elements} for $\mathcal{S}$ and a Multilayer Perception (MLP) for $\bm \phi(\bm{\theta}; \bm{x})$. Specifically, we flatten the input image into a one-dimensional vector before feeding into the $\bm \phi(\bm{\theta}; \bm{x})$, and obtain the generated perturbation by reshaping it back to the original size. Details of the MLP architecture and training hyperparameters can be found in the \textit{Appendix}~\ref{app:implement_details}.

However, we observe that directly optimizing Equation~\ref{eq1} not only targets purified models but also successfully attacks the clean model. We conjecture that this may stem from the inaccurate inversion of reactivated backdoor triggers, which can exploit a clean model in a manner similar to adversarial examples \cite{qin2022boosting,ilyas2019adversarial,liu2016delving}. To mitigate such adversarial behavior, we introduce a regularization term aimed at minimizing backdoor reactivation on the clean model. Given that accessing the clean model may not be practical, we utilize the EP model $\bm{f}(\bm{W}_e;\bm{x})$ as a surrogate model instead.
In sum, we formulate the overall optimization objective as follows:
\begin{equation}
\label{eq2}
    \min_{\bm{\theta}} \Big\{\mathbb{E}_{(\bm{x},\bm{y})\sim\mathcal{D}_c}[\mathcal{S}(\bm{l}(\bm{W}_{ra};\bm{x}), \bm{l}(\bm{W}_p;\epsilon*\bm \phi(\bm{\theta}; \bm{x})+\bm{x}))+\alpha*\mathcal{L}(\bm{f}(\bm{W}_e; \epsilon*\bm{\phi}(\bm{\theta};\bm{x})+\bm{x}), \bm{y})] \Big\},
\end{equation}
where $\alpha$ represents the balance coefficient and the cross-entropy loss is used for $\mathcal{L}$.

We demonstrate our experimental results against five purification methods on CIFAR-10 in Figure~\ref{fig:ba}. Here, we report the \textit{C-ASR} and \textit{P-ASR}, which represent the ASR when evaluating with perturbed clean and perturbed poisoned images, respectively. Notably, our QRA could effectively reactivate the backdoor behaviors on purified models, resulting in a significant increase of $66.13\%$ on average in P-ASR. Our experiments also demonstrate a consistently low C-ASR on purified models, which indicates that the perturbations generated by QRA effectively reactivate the backdoored examples without affecting the predictions of benign images. Besides, the perturbation generated with QRA exclusively works on the output of backdoored samples on purified models, leading to both a low C-ASR and P-ASR on clean models. This observation further indicates that \textit{the reversed pattern generated by QRA is not a typical adversarial perturbation but rather an accurate depiction of the parameter changes necessary for backdoor reactivation.}

\begin{wrapfigure}[18]{r}{0.5\textwidth}
\vspace{-0.4cm}
\centering
\centerline{\includegraphics[width=0.53\textwidth]{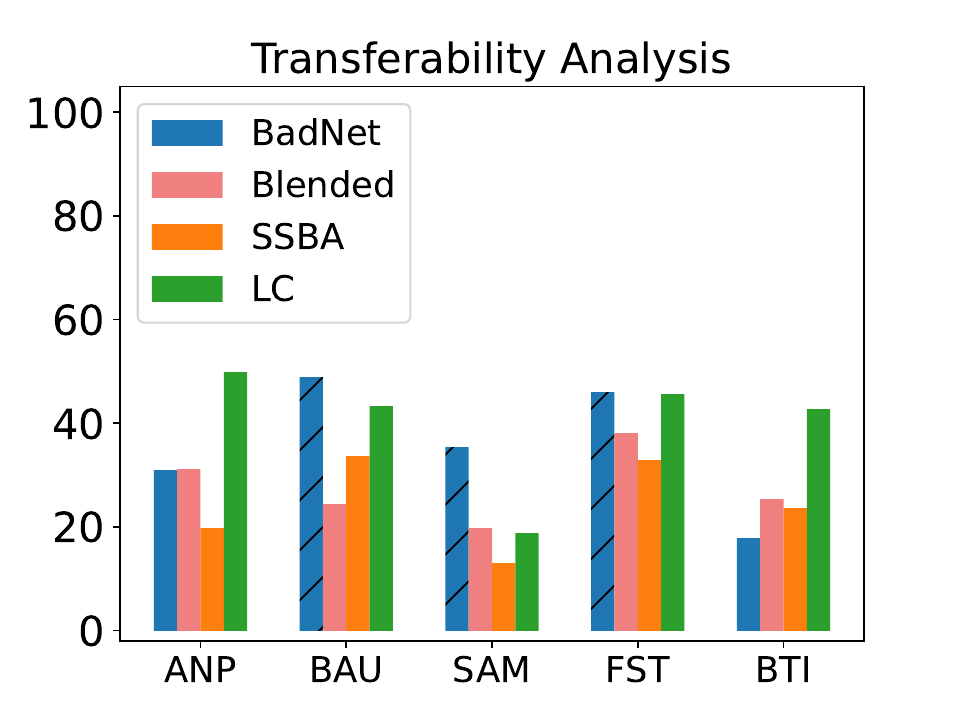}}
\caption{\small The results of the QRA transferability. The defense method used in the attack is represented on the $x$-axis, while the $y$-axis shows the average P-ASR across other purifications.}
\label{fig:transfer}
\end{wrapfigure}
Furthermore, it is worth noting that attackers may lack knowledge about the specific backdoor defense techniques utilized by the defender in practice. Thus, we embark on an initial investigation to explore the transferability of our QRA method across unknown purification methods. Specifically, we aim to determine whether the reversed perturbations optimized for one particular defense method can effectively attack purified models with other purification techniques. As shown in Figure \ref{fig:transfer}, our QRA demonstrates a degree of successful transferability across various defense techniques, achieving an average P-ASR of $32.1\%$ against all purification techniques. These results underscore the potential of QRA to attack purified models, even without prior knowledge of the defense methods employed by defenders. It also highlights the practical application of QRA in real-world situations.

\section{Investigating and Mitigating Superficial Safety}
\subsection{Investigating the Superficial Safety through Linear Mode Connectivity}

While our previous evaluations indicate that only the EP model demonstrates exceptional post-purification robustness compared to current backdoor safety tuning methods, the factors contributing to the effectiveness of EP remain unclear. Motivated by prior studies examining fine-tuned models \cite{garipov2018loss,frankle2020linear,neyshabur2020being}, we propose to investigate this intriguing phenomenon from the perspective of the loss landscape using Linear Mode Connectivity (LMC).

\begin{figure}
\centerline{\includegraphics[width=1.05\textwidth]{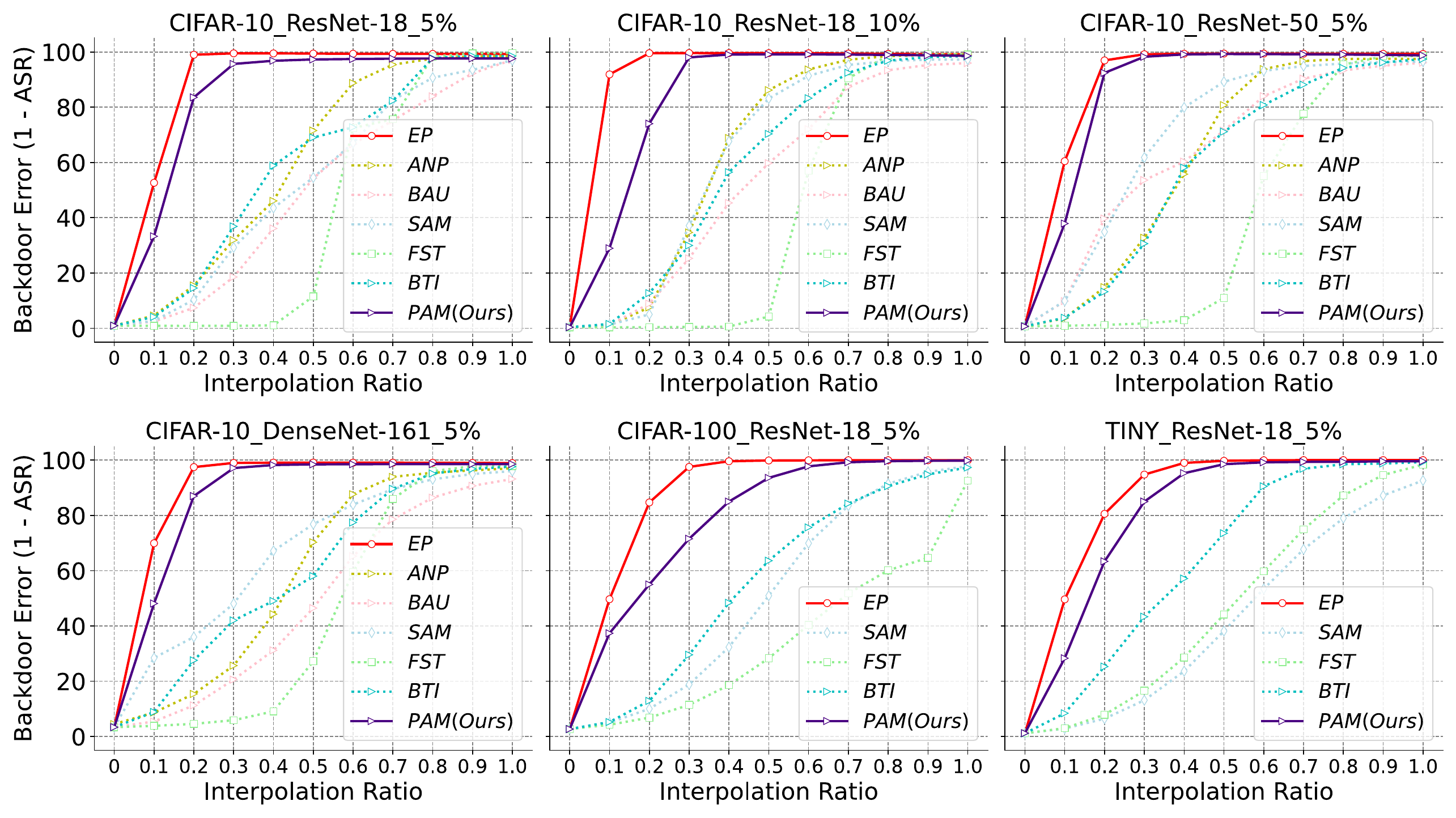}}
    \caption{\small The evaluation of backdoor-connected path against various attack settings. The x-axis and y-axis denote the interpolation ratio $t$ and backdoor error (1-ASR) respectively. For each attack setting, we report the average results among backdoor attacks.}
    \label{fig:lmc}
\end{figure}

\begin{figure}
\centerline{\includegraphics[width=1.05\textwidth]{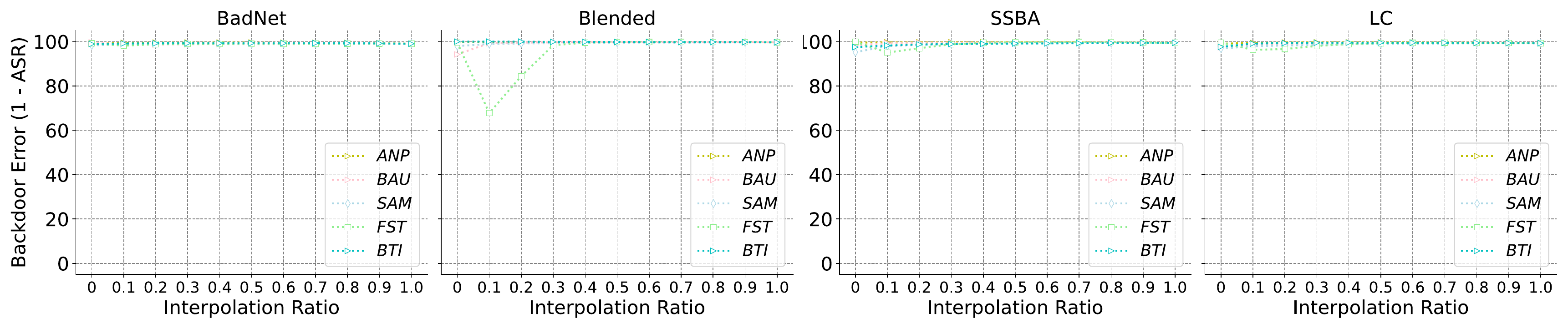}}
    \caption{\small The LMC path connected from other defense techniques to EP. We evaluate the LMC results on CIFAR-10 with ResNet-18, and set the poisoning rate to $5\%$.}
    \label{fig:lmc_connected_to_ep}
\end{figure}

Following~\cite{frankle2020linear,neyshabur2020being}, let $\mathcal{E}(\bm{W};\mathcal{D}_{l})$ represent the testing error of a model $\bm{f}(\bm{W};\bm{x})$ evaluated on a dataset $\mathcal{D}_{l}$. For $\mathcal{D}_{l}$, we use backdoor testing samples. \underline{$\mathcal{E}_{t}(\bm{W}_{0},\bm{W}_{1};\mathcal{D}_{l}) = \mathcal{E}((1-t)\bm{W}_{0}+t\bm{W}_{1};\mathcal{D}_{l})$} for $t\in[0,1]$ is defined as the error path of model created by linear interpolation between the $\bm{f}(\bm{W}_{0};\bm{x})$ and $\bm{f}(\bm{W}_{1};\bm{x})$. We also refer to it as the backdoor-connected path. Here we denote the $\bm{f}(\bm{W}_{0};\cdot)$ as backdoored model and $\bm{f}(\bm{W}_{1};\cdot)$ as the purified model. We show the LMC results of the backdoor error in Figure \ref{fig:lmc} and \ref{fig:lmc_connected_to_ep}. For each attack setting, we report the average results among backdoor attacks. More results on other datasets and models are shown in~\textit{Appendix}~\ref{app:detailed_lmc}.

\label{sec:lmc-analysis} 

\paragraph{The backdoored model and purified models reside in separate loss basins, linked by a backdoor-connected path.} 
We present the results of LMC between purified and backdoored models in Figure~\ref{fig:lmc}. It is clear from the results that all purified models exhibit significant error barriers along the backdoor-connected path to backdoored model. This indicates that backdoored and purified models reside in different loss basins. Additionally, we conduct LMC between purified models with EP and with other defense techniques, as depicted in Figure~\ref{fig:lmc_connected_to_ep}. We observe a consistently high error without barriers, which indicates that these purified models reside within the same loss basin. Based on these two findings, we conclude that backdoored and purified models reside in two distinct loss basins connected through a backdoor-connected path.

\paragraph{EP deviates purified models from the backdoored model along the backdoor-connected path, resulting in a more distantly robust solution.} Although introducing a high loss barrier, we observe notable distinctions between the LMC of the EP model (red solid line) and purified models (dotted lines). We observe a stable high backdoor error along the backdoor-connected path of EP until $t<0.2$, where the interpolated model parameter $\bm{W}$ has over 80\% weight from the \text{backdoored model}. In contrast, other purification models show a tendency to exhibit significant increases in ASR along the path, recovering more than $20\%$ ASR when $t<0.5$, while the ASR for the EP model remains low ($\leq 2\%$).
This clear contrast suggests: \textit{\textbf{1)} the current purification methods prematurely converge to a non-robust solution with low ASR, which is still close to the backdoored model along the backdoor-connected path; \textbf{2)} compared with purified models, EP makes the purified model significantly deviate from the backdoored checkpoint along the backdoor-connected path, resulting in a more robust solution against RA.}

\paragraph{Accurately specified supervision is crucial for achieving stable backdoor safety.} As demonstrated in our observations, the EP method attains stable robustness in the context of the RA, whereas its proxy version, the BTI method, employs reversed backdoor data as a substitute for real backdoor data, resulting in suboptimal post-purification robustness. Furthermore, notable discrepancies are evident in the Backdoor LMC results. These findings underscore that current methods for reversing backdoor triggers are still unable to accurately recover all backdoor features \cite{rando2024competition}, thereby emphasizing the importance of precisely specified supervision in achieving stable backdoor safety. Although data generated by the BTI method does not accurately recover all backdoor features, it could serve as an effective and usable supervision dataset. In the following section, we propose an improved safety tuning method designed to mitigate superficial safety concerns based on this proxy dataset.

\subsection{Enhancing Post-Purification Robustness Through Path-Aware Minimization}
\label{sec:pam}

Motivated by our analysis, we propose a simple tuning defense method called \textbf{Path-Aware Minimization (PAM)}, which aims to enhance post-purification robustness by promoting more deviation from the backdoored model along the backdoor-connected path like the EP method.

Since there are no real backdoor samples $\bm{x}_p$ available, we employ the synthetic backdoored samples $\bm{x}_r$ from the trigger-reversing method BTI~\cite{xu2023towards} as a substitute to get the backdoor-connected path. 
Although BTI has a similar LMC path curve with the EP model in Figure~\ref{fig:lmc_connected_to_ep},  as we have discussed, tuning solely with $\bm{x}_r$ would lead to finding a non-robust solution with low ASR.

To avoid converging to such a solution, we propose utilizing the gradients of an interpolated model $\bm{W}_d$ between $\bm{W}_0$ and $\bm{W}$ to update the current solution $\bm{W}$. As illustrated in Figure~\ref{fig:lmc}, the interpolated model, which lies between $\bm{W}_0$ and $\bm{W}$, exhibits a higher ASR compared to $\bm{W}$. By leveraging the gradients from the interpolated model, we can perform additional updates on the $\bm{W}$ which prevents premature convergence towards local minima and results in a solution that deviates from the backdoored model along this path. Specifically, for $\bm{W}$, we first take a path-aware step $\rho\frac{\bm{W}_{d}}{\|\bm{W}_{d}\|_2}$ ($\bm{W}_{d}=\bm{W}_{0}-\bm{W}$) towards to $\bm{W}_0$ and obtain the interpolated model $\bm{W}+\rho\frac{\bm{W}_{d}}{\|\bm{W}_{d}\|_2}$. Then we compute its gradient on $\bm{x}_r$ to update $\bm{W}$. We formulate our objective function as follows:

\begin{equation}
\label{eq:pam}
    \min_{\bm{W}} \Big\{\mathbb{E}_{(\bm{x},\bm{y})\sim\mathcal{D}_{r}\cup\mathcal{D}_{t}}[\mathcal{L}(\bm{f}(\bm{W}+\rho\frac{\bm{W}_{d}}{\|\bm{W}_{d}\|_2}; \bm{x}), \bm{y})] \Big\},\ \ \ 
    s.t.\   \bm{W}_{d}=\bm{W}_{0}-\bm{W},
\end{equation}

where $\rho$ represents the size of the path-aware step. Typically, a larger $\rho$ indicates a larger step towards the backdoored model $\bm{W}_0$ along our backdoor-connected path and also allows us to obtain a larger gradient update for $\bm{W}$, which results in more deviation from the backdoored model along the backdoor-connected path. The detailed algorithm is summarized in the Algorithm~\ref{alg:pam}.

\begin{table}[t]
\centering
\caption{\small The post-purification robustness performance of PAM on CIFAR-10. The \textit{O-Backdoor} indicates the original performance of backdoor attacks, \textit{O-Robustness} metric represents the purification performance of the defense method, and the \textit{P-Robustness} metric denotes the post robustness after applying RA. All metrics are measured in percentage (\%).
}
\resizebox{\textwidth}{!}{\begin{tabular}{c|c|cc|cc|cc|cc|cc}

\toprule[1pt]
\multirow{2}{*}{Method}  & \multirow{2}{*}{Evaluation Mode} & \multicolumn{2}{c|}{BadNet} & \multicolumn{2}{c|}{Blended} & \multicolumn{2}{c|}{SSBA} & \multicolumn{2}{c|}{LC} & \multicolumn{2}{c}{Avg} \\ \cmidrule{3-12} 
                         &                                 & C-Acc($\uparrow$)            & ASR($\downarrow$)            & C-Acc($\uparrow$)            & ASR($\downarrow$)            & C-Acc($\uparrow$)             & ASR($\downarrow$)            & C-Acc($\uparrow$)          & ASR($\downarrow$) & C-Acc($\uparrow$)          & ASR($\downarrow$)     \\ \midrule
\multirow{5}{*}{ResNet18 (5\%)}  & O-Backdoor &94.04&99.99&94.77&100.0&94.60&96.38&94.86&99.99&94.57&99.09\\
& O-Robustness (EP)    &93.57&0.94&93.47&3.37 &93.39 &0.53 &93.64 &0.52&93.52&1.34 \\
& P-Robustness (EP) &92.27 &1.08 & 93.36 & 4.93 &92.05 & 2.47 &93.69&0.41&92.84&2.22         \\ \cmidrule{2-12} 
& O-Robustness (PAM) &92.11&1.14&93.34&1.67&92.96&1.24&92.32&4.92&92.68&2.24\\
                         & P-Robustness (PAM)  &91.66&3.90&93.38&2.69&92.20&3.31&92.15&8.31&92.35&4.55  \\ \midrule
                         
\multirow{5}{*}{ResNet18 (10\%)} & O-Backdoor &93.73&100.0&94.28&100.0&94.31&98.71&85.80&100.0&92.03&99.68
\\
& O-Robustness (EP)    & 92.78&0.93 &93.34&1.52 &93.10&0.71&92.12&0.16 &92.84&0.83  \\
                         & P-Robustness (EP) & 92.17&1.58 &93.06&4.78 &92.04&3.70&91.79&2.03&92.27&3.02   \\\cmidrule{2-12} 
& O-Robustness (PAM)    &92.43&1.73&92.63&0.22&92.89 &1.37 &91.06 &2.31&92.25&1.41           \\
                         & P-Robustness (PAM) &91.77 &1.47 & 91.85 & 7.44 &92.01 & 2.63 &90.98&3.37 &91.65&3.73        \\ \midrule
\multirow{5}{*}{ResNet50 (5\%)}    & O-Backdoor &93.81&99.92&94.53&100.0&93.65&97.70&94.57&99.90&94.14&99.38\\
& O-Robustness (EP)    &92.84&0.98&92.10&1.12&91.84&0.43&92.91&0.41&92.42&0.74\\
                         & P-Robustness (EP) & 92.33&1.48&91.36&3.51&89.26&6.28&92.33&1.71&91.32&3.25   \\
\cmidrule{2-12} 
& O-Robustness (PAM) & 92.58&0.94&92.59&0.24&92.36&1.62&91.99&2.14&92.38&1.23 \\
                         & P-Robustness (PAM)  &91.49&1.46&92.75&0.64&92.32&14.23&90.60&3.54&91.79&4.97\\ \midrule

\multirow{5}{*}{DenseNet161 (5\%)}    & O-Backdoor &89.85&100.0&89.59&98.72&88.83&86.75&90.13&99.80&89.60&96.32\\
& O-Robustness (EP)   & 88.58&1.52 &88.00&3.21&88.17&0.69&88.59&0.56&88.34&1.50\\
& P-Robustness (EP) &88.03&2.77&87.13&3.54&86.33&1.47&88.94&0.95&87.61&2.18 \\
\cmidrule{2-12} 
 
& O-Robustness (PAM) & 88.70&1.22&87.02&1.08&87.62&1.61&87.70&1.81&87.76&1.43\\
                         & P-Robustness (PAM)  &87.03&3.04&86.49&1.94&85.89&3.72&86.44&9.39&86.42&4.52
                         \\ \bottomrule[1pt]
\end{tabular}}
\label{tab:main_exp_cifar10}
\end{table}

\begin{table}[t]
\centering
\caption{\small The post-purification robustness performance of PAM on CIFAR-100 and Tiny-ImageNet. Note that we omit the LC attack for both the CIFAR-100 and Tiny-ImageNet, as it does not consistently achieve successful backdoor implantation. All metrics are measured in percentage (\%).
}
\resizebox{\textwidth}{!}{\begin{tabular}{c|c|cc|cc|cc|cc}

\toprule[1pt]
\multirow{2}{*}{Method}  & \multirow{2}{*}{Evaluation Mode} & \multicolumn{2}{c|}{BadNet} & \multicolumn{2}{c|}{Blended} & \multicolumn{2}{c|}{SSBA} & \multicolumn{2}{c}{Avg} \\ \cmidrule{3-10} 
                         &                                 & C-Acc($\uparrow$)            & ASR($\downarrow$)            & C-Acc($\uparrow$)            & ASR($\downarrow$)            & C-Acc($\uparrow$)             & ASR($\downarrow$)            & C-Acc($\uparrow$)          & ASR($\downarrow$)    \\ \midrule

\multirow{5}{*}{CIFAR-100 (5\%)}    & O-Backdoor &78.91&99.51&78.98&100.0&78.59&92.38&78.83&97.30\\
& O-Robustness (EP)    & 76.89&0.04&76.95&0.04&76.49&0.05&76.78&0.43\\
& P-Robustness (EP) &76.82&0.10&76.18&0.07&76.12&1.32&76.37&0.50  \\ \cmidrule{2-10} 
& O-Robustness (PAM) &74.97&0.29&76.64&0.19&75.07&0.14&75.56&0.21\\
                         & P-Robustness (PAM) &75.72&0.76&76.12&0.13&74.75&1.95&75.53&0.95
 \\ \midrule
\multirow{5}{*}{TINY (5\%)}    & O-Backdoor &72.60&99.01&73.68&99.99&73.02&97.05&73.10&98.63\\
& O-Robustness (EP)   &70.90&0.02&71.11&0.01&70.24&0.01&70.75&0.13\\
 & P-Robustness (EP) &70.59&0.65&70.93&0.09&69.98&1.90&70.50&0.88  \\ \cmidrule{2-10} 
& O-Robustness (PAM) &68.78&0.06&67.89&0.14&68.01&4.47&68.23&1.56\\
                         & P-Robustness (PAM)
                         &68.97&7.81&66.86&0.37&67.39&14.26&67.74&7.48
                         \\ \bottomrule[1pt]
\end{tabular}}
\label{tab:main_exp_cifar100_tiny}
\end{table}

\paragraph{Post-Purification Robustness of PAM.} 
\label{sec:main_exp}

\begin{wrapfigure}[16]{r}{0.55\textwidth}
\vspace{-0.5cm}
\begin{minipage}{0.55\textwidth}
\small
\begin{algorithm}[H] 
\small
	\caption{Path-Aware Minimization (PAM)} 
	\label{alg:pam}
	\begin{algorithmic}[1]
		\REQUIRE Tuning dataset $\mathcal{D}_{T}=\mathcal{D}_{r} \cup \mathcal{D}_{t}$; Backdoored model $\bm{f}(\bm{W}_0; \bm{x})$; Learning rate $\eta$;  Path-aware step size $\rho$; Tuning iterations $I$ 
		\ENSURE Purified model
            
            \STATE Initialize $\bm{W}_{1}$ with $\bm{W}_0$ 
            \FOR{$i=1, \dots, I$}
            \STATE Sample a mini-batch $\mathcal{B}_i$ from the tuning set $\mathcal{D}_T$;
            \STATE Calculate parameter difference: $\bm{W}_d=\bm{W}_0-\bm{W}_i$;
            \STATE Obtain interpolated parameters: $\Tilde{\bm{W}_i}=\bm{W}_i+\rho\frac{\bm{W}_d}{\|\bm{W}_d\|_2}$;
            \STATE Calculate gradients of $\Tilde{\bm{W}_i}$: \\$\bm{g}_{\Tilde{\bm{W}_i}} = \nabla_{\Tilde{\bm{W}_i}}\frac{1}{|\mathcal{B}_i|}\sum_{(\bm{x},\bm{y})\in\mathcal{B}_i} \mathcal{L}(\bm{f}(\Tilde{\bm{W}_i}; \bm{x}, \bm{y})$
            \STATE Update current parameters: $\bm{W}_{i+1}=\bm{W}_i-\eta\bm{g}_{\Tilde{\bm{W}_i}}$
            \ENDFOR
            \RETURN Purified model $\bm{f}(\bm{W}_{I}; \bm{x})$ 		  
	\end{algorithmic} 
\end{algorithm}
\end{minipage}
\end{wrapfigure}

We evaluate the post-purification robustness of PAM against RA and make a comparison with EP. Using the same experimental settings in Section~\ref{sec:problem-setup}, we set $\rho$ to $0.5$ for Blended and SSBA and $0.9$ for the BadNet and LC attack on CIFAR-10 and set $\rho$ to $0.4$ for both CIFAR-100 and Tiny-ImageNet. The results on CIFAR-10, CIFAR-100 and Tiny-ImageNet are shown in Table~\ref{tab:main_exp_cifar10} and Table~\ref{tab:main_exp_cifar100_tiny}.  We could observe that PAM significantly improves post-purification robustness against RA. It achieves \textit{a comparable robustness performance to the EP}, with an average ASR lower than $4.5\%$ across all three datasets after RA. In comparison to previous experimental results in Section~\ref{sec:retuning-attack}, our PAM outperforms existing defense methods by a large margin in terms of post-purification robustness. Our PAM also achieves a stable purification performance (O-ASR), reducing the ASR below $2\%$ on all three datasets and preserves a high C-Acc as well, yielding only around $2\%$ drop against the original performance of C-Acc.

To further verify the post-purification with PAM, following the experimental setting in Section~\ref{sec:lmc-analysis}, we also show the LMC results of PAM in Figure~\ref{fig:lmc}. It is clearly observed that our PAM significantly deviates purified models from the backdoored model along the backdoor-connected path, leading to a robust solution similar to the EP method. This further confirms our findings about post-purification robustness derived from LMC.

\begin{figure}
\centerline{\includegraphics[width=1.04\textwidth]{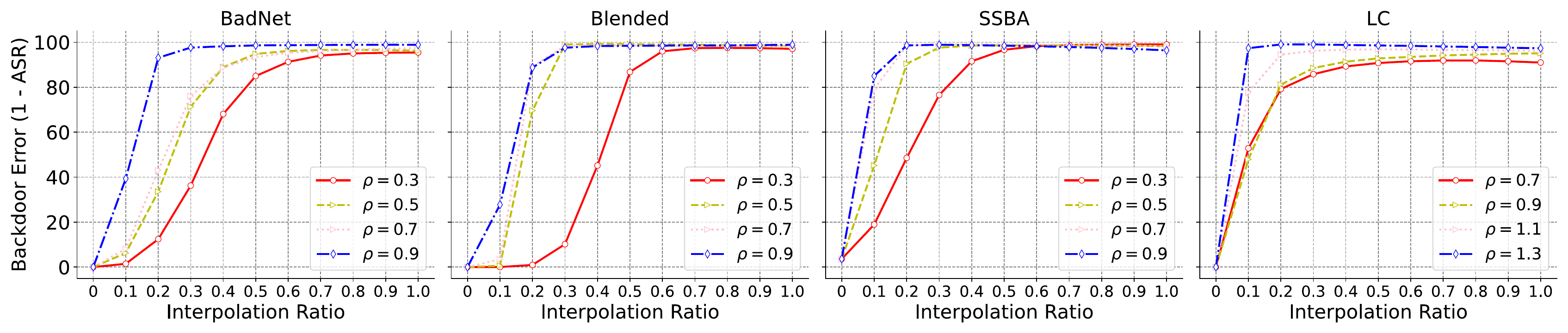}}
    \caption{\small Ablation studies of PAM across different values of $\rho$ against four types of backdoor attacks. We conduct our evaluations on CIFAR-10 with ResNet-18.}
    \label{fig:rho}
\end{figure}

\paragraph{Sensitivity Analysis of $\rho$.}
\label{sec:rho_analysis}
We evaluate the performance of PAM with various values of $\rho$ and conduct experiments on CIFAR-10 with ResNet-18 against four attacks. The experimental results are shown in Figure~\ref{fig:rho}. Note that as $\rho$ increases, we increase the error barrier along the connected path, indicating an enhanced deviation of our purified model from the backdoored model. However, simply increasing $\rho$ would also compromise the competitive accuracy (C-Acc) of the purified model. In practice, it is essential to select an appropriate $\rho$ to achieve a balance between post-purification robustness and C-Acc. We present the model performance across various $\rho$ values in Table \ref{tab:result_rho} of the \textit{Appendix}. We can observe that as $\rho$ rises, there is a slight decrease in clean accuracy alongside a significant enhancement in robustness against the RA. Additionally, we note that performance is relatively insensitive to $\rho$ when it exceeds 0.3. Given that we primarily monitor C-Acc (with the validation set) in practice, we aim to achieve a favorable trade-off between these two metrics. Therefore, we follow the approach of FST \cite{min2024towards} and select $\rho$ to ensure that C-Acc remains above a predefined threshold, such as $92\%$.

\section{Conclusions and Limitations}
\label{sec:conclusion}

In this paper, we seek to address the following question: Do current backdoor safety tuning methods genuinely achieve reliable backdoor safety by merely relying on reduced Attack Success Rates? To investigate this issue, we first employ the Retuning Attack to evaluate the post-purification robustness of purified models. Our primary experiments reveal a significant finding: existing backdoor purification methods consistently exhibit an increased ASR when subjected to the RA, highlighting the superficial safety of these approaches. Building on this insight, we propose a practical Query-based Reactivation Attack, which enables attackers to re-trigger the backdoor from purified models solely through querying. We conduct a deeper analysis of the inherent vulnerabilities against RA using Linear Mode Connectivity, attributing these vulnerabilities to the insufficient deviation of purified models from the backdoored model along the backdoor-connected path. Inspired by our analysis, we introduce a simple tuning defense method, Path-Aware Minimization, which actively promotes deviation from the backdoored model through additional model updates along the interpolated path. Extensive experiments demonstrate the effectiveness of PAM, surpassing existing purification techniques in terms of post-purification robustness.
 
This study represents an initial attempt to evaluate post-purification robustness via RA. While we propose the practical QRA method, future work is essential to develop more efficient evaluation techniques that can faithfully assess post-purification robustness. The need for such evaluations is critical, as they ensure that the perceived safety of purified models is not merely superficial. Additionally, we recognize the significant potential for enhancing QRA in the context of transfer attacks, which we aim to explore in future research. Furthermore, we plan to broaden our research by incorporating additional backdoor attack strategies and safety tuning methods applicable to generative models, such as LLMs and diffusion models \cite{hubinger2024sleeper,qi2023fine,anwar2024foundational,rando2024competition}, in future work. We will also apply our existing framework of evaluation, analysis, and safety tuning method to research on unlearning in large models.

{
\clearpage
\small
\bibliographystyle{plainnat}
\bibliography{reference}
}







\clearpage
\appendix

\section{Social Impact}
\label{app:impact}

The prevalence of Deep Neural Networks (DNNs) in modern society relies heavily on massive amounts of training data from diverse sources. However, in the absence of rigorous monitoring mechanisms, these data resources become susceptible to malicious manipulation, resulting in unforeseen and potentially harmful consequences. Among the various concerns associated with the training dataset, backdoor attacks pose a significant threat. These attacks can manipulate the behavior of a well-trained model by poisoning the training set with backdoored samples, often at a low cost and without requiring complete control over the training process. While existing defense methods have demonstrated effective backdoor purification by achieving low Attack Success Rates (ASR), they still exhibit vulnerabilities that allow adversaries to reactivate the injected backdoor behavior easily. In our work, instead of solely focusing on backdoor ASR, we investigate the effectiveness of modern purification techniques from the perspective of post-purification robustness. We aim to enhance the post-purification robustness of backdoor defense, mitigating the potential for malicious manipulation of deployed models even after backdoor purification. In sum, our work hopes to move an initial step towards improving post-purification robustness while also contributing to another aspect of understanding and enhancing machine learning security.

\section{Experimental Settings}
\label{app:exp_settings}
In this section, we provide detailed information about the experimental settings used in our evaluations. This includes the dataset, training details, and the selection of hyperparameters. All experiments were conducted using 4 NVIDIA 3090 GPUs. We ran all experiments 3 times and averaged all results over 3 random seeds.

\subsection{Datasets and Models}
We follow previous studies~\cite{wu2022backdoorbench,li2021anti,min2024towards,wu2021adversarial} on backdoor learning, and conduct our experiments on three widely used datasets including CIFAR-10, CIFAR-100, and Tiny-ImageNet.

\begin{itemize}
  \item CIFAR-10 is a widely used dataset in the backdoor literature, comprising images with a resolution of $32\times32$ and 10 categories. For backdoor training, we utilize the ResNet-18 model for main evaluation, a commonly used architecture in previous studies~\cite {xu2023towards,min2024towards,zhu2023enhancing}. Additionally, we explore other architectures, including the ResNet-50 and DenseNet-161.
  \item CIFAR-100 and Tiny-ImageNet are two large-scale datasets compared to the CIFAR-10, which include 100 and 200 different categories, respectively. Similar to previous work~\cite{xu2023towards,min2024towards}, we utilize the pretrained ResNet-18 on ImageNet-1K provided by \textit{PyTorch} to implement backdoor attacks since directly training from scratch would result in an inferior model performance on C-Acc, hence is not practical in real-world scenarios. 
\end{itemize}
\subsection{Implementation Details}
\label{app:implement_details}
\textbf{Attack Configurations} We implement $4$ representative poisoning-based attacks and generally follow the implementation from the BackdoorBench\footnote{\url{https://github.com/SCLBD/backdoorbench}}. For the BadNet, we utilize the $3\times3$ checkerboard patch as triggers and choose the lower right corner of the image for backdoor injection by default; for the Blended, we adopt the Gaussian Noise as the backdoor trigger. We set the blend ratio to $0.1$ for backdoor training and increase the blend ratio to $0.2$ during the inference phase; for SSBA and LC, we follow the original implementation from BackdoorBench without making modifications. In our implementation, we set the default poisoning rate to $5\%$, which is commonly used in previous studies~\cite{min2024towards,zhu2023enhancing} and additionally explore various poisoning rates including both $1\%$ and $10\%$. Note that we do not adopt a lower poisoning rate since most of the methods suffer from effectively removing backdoor effects when the poisoning rate is extremely low as indicated by~\cite{min2024towards}. For all backdoor attacks, the target label is set to be $0$ by default.

For CIFAR-10, we adopt an initial learning rate of $0.1$ to train all the backdoored models for $100$ epochs. For both the CIFAR-100 and Tiny-ImageNet, we utilize pretrained backbones and initialize the classifiers with appropriate class numbers. We adopt a smaller learning rate of $0.001$ and fine-tune the models for $10$ epochs. We upscale the image size up to $224*224$ during both the training and inference stages following the implementation of~\cite{min2024towards}.

\textbf{Baseline Defense Configurations}
We evaluate several mainstream purification techniques including pruning-based defense (ANP), tuning-based defenses (I-BAU, FT-SAM, and FST), and the state-of-the-art trigger-inversing strategy BTI.

\begin{itemize}
\item EU: We add real backdoor triggers to $10\%$ of the overall benign tuning samples, and keep their ground-truth labels unchanged. We fine-tune the model on our tuning dataset for $20$ epochs.
\item ANP: We follow the original implementation in BackdoorBench, using the default hyperparameters. Regarding model pruning, we set the threshold range from $0.4$ to $0.9$ and report the best purification result with a low ASR and a high model performance on C-Acc.

\item I-BAU: We follow the implementation in BackdoorBench with default configurations. We set the fixed-point approximation iterations to $5$ and fine-tune backdoored models for $20$ epochs.

\item FT-SAM: We follow the implementation from~\cite{min2024towards} and set the neighborhood size to $1$ for CIFAR-10 and $0.5$ for both the CIFAR-100 and Tiny-ImageNet. We set the initial learning rate to $0.01$ and decrease the learning rate to $0.001$ for both the CIFAR-100 and Tiny-ImageNet. We fine-tune backdoored models for $20$ epochs on all datasets.

\item FST: For FST, we follow the original implementation\footnote{\url{https://github.com/AISafety-HKUST/stable_backdoor_purification}}. We set the feature-shift parameter to $0.1$ and the learning rate to $0.01$ on CIFAR-10. On the CIFAR-100 and Tiny-ImageNet, we decrease the feature-shift parameter to $0.001$ and adopt an initial learning rate of $0.005$ for better C-Acc.

\item BTI: We follow its original implementation\footnote{\url{https://github.com/xuxiong0214/BTIDBF}} and adopt the BTI (U) since it achieves a better purification performance. We adopt the default hyperparameters for trigger inversion including $20$ iterations for decoupling benign features, and $30$ iterations for training backdoor generators, and set the norm bound to $0.3$ by default. We perform BTI (U) until the loss converges.

\item PAM: On CIFAR-10, we set the path-aware step size to $0.5$ for the Blended and SSBA; while we increase the step size to $0.9$ for both the BadNet and the LC attack. We use a step size of $0.4$ on CIFAR-100 and Tiny-ImageNet to maintain the model performance on C-Acc.

\end{itemize}

\textbf{RA Configurations}
In this section, we provide the detailed experimental setting of the RA in our revisiting Section~\ref{sec:revisiting-evaluations}. For the CIFAR-10, we adopt $5$ samples for BadNet, $1$ samples for Blended, $10$ samples for SSBA, and $2$ samples for LC attack to perform RA. For both the CIFAR-100 and Tiny-ImageNet, we increase the number of poisoned images to $25$ samples for BadNet, $10$ samples for Blended, and $35$ samples for SSBA to perform RA. We set the learning rate to $0.01$ for the CIFAR-10 and $0.001$ for the CIFAR-100 and Tiny-ImageNet to maintain C-Acc. However, directly fine-tuning with these poisoned samples would negatively impact the performance of C-Acc. Therefore, in addition to the backdoored samples, we introduce extra benign samples during the fine-tuning process. We fix the size of the tuning images to $2\%$ of the overall training dataset, which are $1000$ samples for CIFAR-10 and CIFAR-100, and $2000$ samples for Tiny-ImageNet. 

Note that there are primarily two reasons for selecting only a limited number of poisoned examples for fine-tuning during our experiments. Firstly, using a larger number of poisoned samples, such as $10$ BadNet samples on CIFAR-10, would significantly increase the ASR of a clean model. This would undermine the reliability of our RA evaluation conducted on the purified models. Secondly, by utilizing only a few poisoned examples, we intentionally expose the purified models to a potential compromise, even if they exhibit a seemingly low ASR. This approach effectively highlights the vulnerability of these purified models, emphasizing their susceptibility to attacks despite achieving a seemingly low ASR.

\textbf{QRA Configurations}
We employ a three-layer Multilayer Perceptron (MLP) to generate the reversed perturbation in the input space. For simplicity, the number of neurons in all internal layers is fixed at $1024$, followed by the Rectified Linear Unit (ReLU) activation function. To train the generator, we use $500$ benign examples and $500$ backdoored examples from the CIFAR-10 dataset. Our training process is conducted over $50$ epochs with a constant learning rate of $0.1$. We set the $\alpha$ to $0.1$ for the LC attack and $0.2$ for the others to achieve a high P-ASR on purified models while simultaneously reducing the attack performance on clean models. During the training process of the perturbation generator, we begin by flattening the input image into a one-dimensional vector before feeding it to the generator. Once we obtain the output from the generator, we reshape it back to the original input size. We then multiply it with the pre-defined budget $\epsilon$ before we integrate it into the images. In our implementation, we fix the $\epsilon$ to $16/255$ for all experiments.

\section{Additional Experiments}
\subsection{Additional Results of RA}
\label{app:detailed_ra}
\textbf{Detailed RA performance}
Our previous experiments in Section~\ref{sec:revisiting-evaluations} only provide the average metrics on CIFAR-10. Therefore, in this section, we provide detailed RA results against each backdoor attack. Specifically, we demonstrate the detailed experimental results of ResNet-18 with $5\%$ poisoning rate in Table~\ref{tab:ra_res18_cifar10_5}; ResNet-18 with $10\%$ poisoning rate in Table~\ref{tab:ra_res18_cifar10_10}; ResNet-50 with $5\%$ poisoning rate in Table~\ref{tab:ra_res50_cifar10_5} and DenseNet-161 with $5\%$ poisoning rate in Table~\ref{tab:ra_dense_cifar10_5}, respectively. Based on these experimental results, our PAM demonstrates effective post-purification robustness against diverse architectures and poisoning rates, which leads to only a small ASR increase (less than $3\%$ on average) after performing RA.

\begin{table}[t]
\centering
\caption{\small The post-purification robustness performance against diverse defense methods. We evaluate the performance on CIFAR-10 with ResNet-18 and set the overall poisoning rate to $5\%$. The \textit{O-Backdoor} indicates the original performance of backdoor attacks, \textit{O-Robustness} metric represents the purification performance of the defense method, and the \textit{P-Robustness} metric denotes the post robustness after applying RA. All metrics are measured in percentage (\%).
}

\resizebox{\textwidth}{!}{\begin{tabular}{c|c|cc|cc|cc|cc}

\toprule[1pt]
\multirow{2}{*}{Method}  & \multirow{2}{*}{Mode} & \multicolumn{2}{c|}{BadNet} & \multicolumn{2}{c|}{Blended} & \multicolumn{2}{c|}{SSBA} & \multicolumn{2}{c}{LC}  \\ \cmidrule{3-10} 
                         &                                 & C-Acc($\uparrow$)            & ASR($\downarrow$)            & C-Acc($\uparrow$)            & ASR($\downarrow$)            & C-Acc($\uparrow$)             & ASR($\downarrow$)            & C-Acc($\uparrow$)          & ASR($\downarrow$)      \\ \midrule
Attack  & O-Backdoor &94.04&99.99&94.77&100.0&94.60&96.38&94.86&99.99\\
\midrule
\multirow{2}{*}{Clean}  & O-Robustness & 95.07 & 0.57 & 95.07 & 0.90 & 95.07 & 0.61  &95.07&0.81       \\
                         & P-Robustness & 94.37 & 0.20 &95.02 &8.12 &94.29 &1.26 &94.93&1.33                \\ \midrule
\multirow{2}{*}{EP} & O-Robustness    &93.57&0.94&93.47&3.37 &93.39 &0.53 &93.64 &0.52           \\
                         & P-Robustness &92.27 &1.08 & 93.36 & 4.93 &92.05 & 2.47 &93.69&0.41         \\ \midrule
                          \midrule
\multirow{2}{*}{ANP}    & O-Robustness &93.75&4.24&94.56&0.30&93.22&0.44&93.40&0.92    \\
                         & P-Robustness  &93.76&100&94.42&54.93&93.99&82.74&94.26&97.92
                         \\ \midrule
\multirow{2}{*}{BAU}    & O-Robustness  &92.63&1.61&93.47&5.77&93.41&1.63&92.80&2.49  \\
                         & P-Robustness  &93.50&100&94.17&100&93.93&93.99&94.38&99.29  \\ \midrule
\multirow{2}{*}{SAM}    & O-Robustness &92.85 &2.14 &93.14 & 2.01 & 93.27 & 4.80 &93.00&3.14   \\
                         & P-Robustness  &92.89 &100 &93.05&78.93&92.98&72.20 &92.57&97.91
                         \\ \midrule

\multirow{2}{*}{FST}    & O-Robustness  &93.89&0.78&93.92&0.81&94.01&0.63&94.09&0.41    \\
                         & P-Robustness   &93.92&100&93.88&99.87&93.91&85.53&94.16&90.36
                         \\ \midrule

\multirow{2}{*}{BTI}    & O-Robustness &92.56&1.08&92.48&0.00&92.21&2.46&93.28&2.64 \\
                         & P-Robustness  &91.15&100&91.86&99.8&91.18&68.57&92.89&99.97
\\ \midrule
\multirow{2}{*}{PAM (Ours)}    & O-Robustness &92.11&1.14&93.34&1.67&92.96&1.24&92.32&4.92
\\
& P-Robustness &\cellcolor{gray!20}91.66&\cellcolor{gray!20}3.90&\cellcolor{gray!20}93.38&\cellcolor{gray!20}2.69&\cellcolor{gray!20}92.20&\cellcolor{gray!20}3.31&\cellcolor{gray!20}92.15&\cellcolor{gray!20}8.31
                         \\ \bottomrule[1pt]
\end{tabular}}

\label{tab:ra_res18_cifar10_5}
\end{table}

\begin{table}[t]
\centering
\caption{\small The post-purification robustness performance against diverse defense methods. We evaluate the performance on CIFAR-10 with ResNet-18 and set the overall poisoning rate to $10\%$. The \textit{O-Backdoor} indicates the original performance of backdoor attacks, \textit{O-Robustness} metric represents the purification performance of the defense method, and the \textit{P-Robustness} metric denotes the post robustness after applying RA. All metrics are measured in percentage (\%).
}
\resizebox{\textwidth}{!}{\begin{tabular}{c|c|cc|cc|cc|cc}

\toprule[1pt]
\multirow{2}{*}{Method}  & \multirow{2}{*}{Mode} & \multicolumn{2}{c|}{BadNet} & \multicolumn{2}{c|}{Blended} & \multicolumn{2}{c|}{SSBA} & \multicolumn{2}{c}{LC}  \\ \cmidrule{3-10} 
                         &                                 & C-Acc($\uparrow$)            & ASR($\downarrow$)            & C-Acc($\uparrow$)            & ASR($\downarrow$)            & C-Acc($\uparrow$)             & ASR($\downarrow$)            & C-Acc($\uparrow$)          & ASR($\downarrow$)      \\ \midrule
Attack  & O-Backdoor &93.73&100.0&94.28&100.0&94.31&98.71&85.80&100.0\\
\midrule
\multirow{2}{*}{Clean}  & O-Robustness &95.07&0.57 & 95.07 & 0.90 & 95.07 & 0.61  &95.07&0.81       \\
                         & P-Robustness & 94.37 & 0.20 &95.02 &8.12 &94.29 &1.26 &94.93&1.33                \\ \midrule
\multirow{2}{*}{EP} & O-Robustness    & 92.78&0.93 &93.34&1.52 &93.10&0.71&92.12&0.16  \\
                         & P-Robustness & 92.17&1.58 &93.06&4.78 &92.04&3.70&91.79&2.03   \\ \midrule\midrule
\multirow{2}{*}{ANP}    & O-Robustness &93.29&0.60 &94.20&0.07 &93.90&2.14&84.72&0.00  \\
                         & P-Robustness  &92.74&100.0&93.77&76.96&93.42&91.92&92.00&91.41
                         \\ \midrule
\multirow{2}{*}{BAU}    & O-Robustness  &91.85&0.82&92.76&5.20 &91.99&6.20&91.80&3.89\\
                         & P-Robustness  & 92.96&100.0&93.19&99.99&93.38&97.32&91.87&99.49\\ \midrule
\multirow{2}{*}{SAM}    & O-Robustness &92.63&1.50&92.64&0.79&91.70&4.01&91.81&4.78 \\
                         & P-Robustness  &92.40&100&92.73&32.16&91.39&81.59&91.73&98.91
                         \\ \midrule

\multirow{2}{*}{FST}    & O-Robustness  &93.41&0.07&93.81&0.08&93.86&0.08&92.43&3.44\\
                         & P-Robustness   &93.38&100.0&93.77&99.96&93.53&93.91&92.27&73.68
                         \\ \midrule

\multirow{2}{*}{BTI}    & O-Robustness &92.36&1.69 &93.17&0.67&93.16&2.09&91.84&1.59\\
                         & P-Robustness  &91.14&100&92.44&83.43&92.48&93.82&91.94&91.93
\\ \midrule
\multirow{2}{*}{PAM (Ours)}    & O-Robustness &92.43&1.73&92.63&0.22&92.89&1.37&91.06&2.31
\\
                         & P-Robustness  &\cellcolor{gray!20}91.77&\cellcolor{gray!20}1.47&\cellcolor{gray!20}91.85&\cellcolor{gray!20}7.44&\cellcolor{gray!20}92.01&\cellcolor{gray!20}2.63&\cellcolor{gray!20}90.98&\cellcolor{gray!20}3.37
                         \\ \bottomrule[1pt]
\end{tabular}}
\label{tab:ra_res18_cifar10_10}
\end{table}

\begin{table}[t]
\centering
\caption{\small The post-purification robustness performance against diverse defense methods. We evaluate the performance on CIFAR-10 with ResNet-50 and set the overall poisoning rate to $5\%$. The \textit{O-Backdoor} indicates the original performance of backdoor attacks, \textit{O-Robustness} metric represents the purification performance of the defense method, and the \textit{P-Robustness} metric denotes the post robustness after applying RA. All metrics are measured in percentage (\%).
}
\resizebox{\textwidth}{!}{\begin{tabular}{c|c|cc|cc|cc|cc}

\toprule[1pt]
\multirow{2}{*}{Method}  & \multirow{2}{*}{Mode} & \multicolumn{2}{c|}{BadNet} & \multicolumn{2}{c|}{Blended} & \multicolumn{2}{c|}{SSBA} & \multicolumn{2}{c}{LC}  \\ \cmidrule{3-10} 
                         &                                 & C-Acc($\uparrow$)            & ASR($\downarrow$)            & C-Acc($\uparrow$)            & ASR($\downarrow$)            & C-Acc($\uparrow$)             & ASR($\downarrow$)            & C-Acc($\uparrow$)          & ASR($\downarrow$)      \\ \midrule
Attack  & O-Backdoor &93.81&99.92&94.53&100.0&93.65&97.70&94.57&99.90\\
\midrule
\multirow{2}{*}{Clean}  & O-Robustness &94.6&0.72 &94.6&0.19 &94.6&0.77&94.60&0.80\\
                         & P-Robustness & 94.06&4.10 &94.25&3.58&93.00&3.03&94.29&4.44                \\ \midrule
\multirow{2}{*}{EP} & O-Robustness    &92.84&0.98&92.10&1.12&91.84&0.43&92.91&0.41\\
                         & P-Robustness & 92.33&1.48&91.36&3.51&89.26&6.28&92.33&1.71   \\ \midrule\midrule
\multirow{2}{*}{ANP}    & O-Robustness &92.26&1.01 &94.51&0.30&92.46&2.08 &92.05&6.32\\
                         & P-Robustness  &93.44&92.02&94.43&76.51&93.01&85.51  &93.67&81.14                      \\ \midrule
\multirow{2}{*}{BAU}    & O-Robustness  &92.17&1.23&92.88&3.82&90.57&6.64&92.63&3.91\\
                         & P-Robustness  &92.85&17.94&93.7&98.49&92.28&92.94&93.14&23.73\\ \midrule
\multirow{2}{*}{SAM}    & O-Robustness &91.53&2.69&92.27&4.86&91.77&2.56&91.82&3.73\\
                         & P-Robustness  &91.39&66.87&92.28&69.04&91.43
&75.49 &91.83&17.96

                         \\ \midrule

\multirow{2}{*}{FST}    & O-Robustness  &92.46&0.38&93.79&0.18&93.09&3.52&92.83&1.62\\
                         & P-Robustness   &93.12&85.08&93.53&99.37&92.81&81.04&92.97&56.07
                         \\ \midrule

\multirow{2}{*}{BTI}    & O-Robustness &92.27&4.67&92.15&0.94&92.12&1.39&92.42&3.20\\
                         & P-Robustness  &92.37&100.0&91.74&99.70&92.05&68.63&91.77&89.93
\\ \midrule
\multirow{2}{*}{PAM (Ours)}    & O-Robustness &92.58&0.94&92.59&0.24&92.36&1.62&91.99&2.14\\
                         & P-Robustness  &\cellcolor{gray!20}91.49&\cellcolor{gray!20}1.46&\cellcolor{gray!20}92.75&\cellcolor{gray!20}0.64&\cellcolor{gray!20}92.32&\cellcolor{gray!20}14.23&\cellcolor{gray!20}90.68&\cellcolor{gray!20}3.54
                         \\ \bottomrule[1pt]
\end{tabular}}
\label{tab:ra_res50_cifar10_5}
\end{table}

\begin{table}[t]
\centering
\caption{\small The post-purification robustness performance against diverse defense methods. We evaluate the performance on CIFAR-10 with DenseNet-161 and set the overall poisoning rate to $5\%$. We omit the performance of ANP against BadNet since it can not achieve successful backdoor purification. The \textit{O-Backdoor} indicates the original performance of backdoor attacks, \textit{O-Robustness} metric represents the purification performance of the defense method, and the \textit{P-Robustness} metric denotes the post robustness after applying RA. All metrics are measured in percentage (\%).
}
\resizebox{\textwidth}{!}{\begin{tabular}{c|c|cc|cc|cc|cc}

\toprule[1pt]
\multirow{2}{*}{Method}  & \multirow{2}{*}{Mode} & \multicolumn{2}{c|}{BadNet} & \multicolumn{2}{c|}{Blended} & \multicolumn{2}{c|}{SSBA} & \multicolumn{2}{c}{LC}  \\ \cmidrule{3-10} 
                         &                                 & C-Acc($\uparrow$)            & ASR($\downarrow$)            & C-Acc($\uparrow$)            & ASR($\downarrow$)            & C-Acc($\uparrow$)             & ASR($\downarrow$)            & C-Acc($\uparrow$)          & ASR($\downarrow$)      \\ \midrule
Attack  & O-Backdoor &89.85&100.0&89.59&98.72&88.83&86.75&90.13&99.80\\
\midrule
\multirow{2}{*}{Clean}  & O-Robustness &90.17&1.27&90.17&1.14&90.17&1.44&90.17&2.17\\
                         & P-Robustness & 89.54&2.11&89.47&1.27 &88.24&4.64 &89.41&4.44          \\ \midrule
\multirow{2}{*}{EP} & O-Robustness    & 88.58&1.52 &88.00&3.21&88.17&0.69&88.59&0.56\\
                         & P-Robustness &88.03&2.77&87.13&3.54&86.33&1.47&88.94&0.95 \\ \midrule\midrule
\multirow{2}{*}{ANP}    & O-Robustness &- &- &89.18&4.56&88.64&2.74&90.04&1.58\\
                         & P-Robustness  &-&-&89.06&93.72&88.65&47.47&89.45&71.10
                         \\ \midrule
\multirow{2}{*}{BAU}    & O-Robustness  &88.00&1.81&87.62&5.43&86.91&17.34&88.02&2.63\\
                         & P-Robustness  &88.52&91.40&88.75&90.40&87.96&78.77&88.84&82.47 \\ \midrule
\multirow{2}{*}{SAM}    & O-Robustness &86.91&2.00&86.96&7.38&86.64&3.92&87.79&2.52\\
                         & P-Robustness  &86.79&100.0&87.09&78.21&86.98&35.97&87.30&12.23
                         \\ \midrule

\multirow{2}{*}{FST}    & O-Robustness  &88.33&5.37&88.56&0.18&88.32&1.63&87.40&1.57\\
                         & P-Robustness   &88.16&99.86&87.92&83.06&87.52&68.82&87.47&72.61
                         \\ \midrule

\multirow{2}{*}{BTI}    & O-Robustness &89.01&0.77&88.22&1.80&88.30&0.53&87.93&2.71\\
                         & P-Robustness  &88.11&97.73&88.35&26.20&87.43&48.47&87.67&100.0
\\ \midrule
\multirow{2}{*}{PAM (Ours)}    & O-Robustness & 88.70&1.22&87.02&1.08&87.62&1.61&87.70&1.81\\
                         & P-Robustness  &\cellcolor{gray!20}87.03&\cellcolor{gray!20}3.04&\cellcolor{gray!20}86.49&\cellcolor{gray!20}1.94&\cellcolor{gray!20}85.89&\cellcolor{gray!20}3.72&\cellcolor{gray!20}86.44&\cellcolor{gray!20}9.39
                         \\ \bottomrule[1pt]
\end{tabular}}
\label{tab:ra_dense_cifar10_5}
\end{table}

\textbf{Evaluation of RA under Lower Poisoning Rate} In addition to the $5\%$ and $10\%$ poisoning rates used in our previous evaluation, we also include a lower poisoning rate of $1\%$ to assess the performance of RA. Specifically, we experiment on CIFAR-10 and utilize the ResNet-18 for a fair comparison. We exclude the LC attack from our experiments since most defense methods struggle to adequately purify the backdoor behavior. As shown in Table~\ref{tab:ra_res18_cifar10_1}, our PAM achieves a tiny increase of the ASR after RA compared to other purification techniques. This further demonstrates the effectiveness of our PAM method in enhancing post-purification robustness.

\begin{table}[t]
\centering
\caption{\small The post-purification robustness performance against diverse defense methods. We evaluate the performance on CIFAR-10 with ResNet-18 and set a lower poisoning rate to $1\%$. We omit the performance of SAM against Blended since it can not achieve successful backdoor purification. The \textit{O-Backdoor} indicates the original performance of backdoor attacks, \textit{O-Robustness} metric represents the purification performance of the defense method, and the \textit{P-Robustness} metric denotes the post robustness after applying RA. All metrics are measured in percentage (\%).
}
\resizebox{0.8\textwidth}{!}{\begin{tabular}{c|c|cc|cc|cc}

\toprule[1pt]
\multirow{2}{*}{Method}  & \multirow{2}{*}{Mode} & \multicolumn{2}{c|}{BadNet} & \multicolumn{2}{c|}{Blended} & \multicolumn{2}{c}{SSBA}  \\ \cmidrule{3-8} 
                         &                                 & C-Acc($\uparrow$)            & ASR($\downarrow$)            & C-Acc($\uparrow$)            & ASR($\downarrow$)            & C-Acc($\uparrow$)             & ASR($\downarrow$)              \\ \midrule
Attack  & O-Backdoor &94.77&100.0&94.90&99.12&94.94&79.98\\
\midrule
\multirow{2}{*}{Clean}  & O-Robustness & 95.07&0.57&95.07&0.90&95.07&0.61\\
                         & P-Robustness & 94.63&1.50&94.90&2.77&94.47&0.98                 \\ \midrule
\multirow{2}{*}{EP} & O-Robustness    & 94.36&0.54 &93.22&0.52&94.16&0.95\\
                         & P-Robustness & 93.36&0.68 &93.36&0.51&92.86&0.80 \\ \midrule\midrule
\multirow{2}{*}{ANP}    & O-Robustness &93.41&0.60&93.84&0.29&94.61&1.13\\
                         & P-Robustness  &94.32&87.11&94.49&0.82&94.37&48.21 
                         \\ \midrule
\multirow{2}{*}{BAU}    & O-Robustness  &92.64&6.89&92.64&3.67&93.18&2.81\\
                         & P-Robustness  &93.88&95.67&93.92&99.91&94.14&58.04\\ \midrule
\multirow{2}{*}{SAM}    & O-Robustness &93.45&2.80&-&-&92.68&3.77 \\
                         & P-Robustness  &93.64&41.72&-&-&92.57&55.51
                         \\ \midrule

\multirow{2}{*}{FST}    & O-Robustness  &94.47&1.48&94.09&0.00&94.20&0.13\\
                         & P-Robustness   &93.53&90.07&93.90&99.90&93.86&80.23
                         \\ \midrule

\multirow{2}{*}{BTI}    & O-Robustness &92.89&1.44&92.82&1.06&92.90&1.52\\
                         & P-Robustness  &92.33&99.98&92.50&95.21&91.92&36.84
\\ \midrule
\multirow{2}{*}{PAM (Ours)}    & O-Robustness &92.52&2.91&92.6&0.10&92.71&1.91\\
                         & P-Robustness  &\cellcolor{gray!20}92.61&\cellcolor{gray!20}1.74&\cellcolor{gray!20}92.91&\cellcolor{gray!20}0.47&\cellcolor{gray!20}91.91&\cellcolor{gray!20}3.45
                         \\ \bottomrule[1pt]
\end{tabular}}
\label{tab:ra_res18_cifar10_1}
\end{table}

\textbf{Evaluation of RA under Other Datasets} In this section, we provide additional evaluations on RA against two other datasets. We present our experimental results in Table~\ref{tab:ra_res18_cifar100_5} and Table~\ref{tab:ra_res18_tiny_5} for the CIFAR-100 and Tiny-ImageNet, respectively. It is worth noting that our PAM method remains effective when applied to these two datasets, resulting in only a minor increase in ASR.

\begin{table}[t]
\centering
\caption{\small The post-purification robustness performance against diverse defense methods. We evaluate the performance on CIFAR-100 with the pretrained ResNet-18 and set the poisoning rate to $5\%$. The \textit{O-Backdoor} indicates the original performance of backdoor attacks, \textit{O-Robustness} metric represents the purification performance of the defense method, and the \textit{P-Robustness} metric denotes the post robustness after applying RA. All metrics are measured in percentage (\%).
}
\resizebox{0.8\textwidth}{!}{\begin{tabular}{c|c|cc|cc|cc}

\toprule[1pt]
\multirow{2}{*}{Method}  & \multirow{2}{*}{Mode} & \multicolumn{2}{c|}{BadNet} & \multicolumn{2}{c|}{Blended} & \multicolumn{2}{c}{SSBA}  \\ \cmidrule{3-8} 
                         &                                 & C-Acc($\uparrow$)            & ASR($\downarrow$)            & C-Acc($\uparrow$)            & ASR($\downarrow$)            & C-Acc($\uparrow$)             & ASR($\downarrow$)             \\ \midrule
Attack  & O-Backdoor &78.91&99.51&78.98&100.0&78.59&92.38\\
\midrule
\multirow{2}{*}{Clean}  & O-Robustness &79.70&0.04&79.70&0.03&79.70&0.04\\
                         & P-Robustness & 78.92&0.25&79.03&0.39&78.30&1.91
                         \\ \midrule
\multirow{2}{*}{EP} & O-Robustness    & 76.89&0.04&76.95&0.04&76.49&0.05\\
                         & P-Robustness &76.82&0.10&76.18&0.07&76.12&1.32  \\ \midrule\midrule

\multirow{2}{*}{SAM}    & O-Robustness &76.27&2.85&76.54&0.44&76.34&3.30\\
                         & P-Robustness  &76.63&95.30&77.16&97.90&75.50&78.35
                         \\ \midrule

\multirow{2}{*}{FST}    & O-Robustness  &73.02&1.71&72.41&0.30&73.54&0.18\\
                         & P-Robustness   &72.93&93.71&72.28&100.0&72.06&59.90
                         \\ \midrule

\multirow{2}{*}{BTI}    & O-Robustness &75.55&5.79&75.68&0.49&75.59&1.77\\
                         & P-Robustness  &75.78&59.29&76.23&70.44&75.07&46.10
\\ \midrule
\multirow{2}{*}{PAM (Ours)}    & O-Robustness &74.97&0.29&76.64&0.19&75.07&0.14\\
                         & P-Robustness  &\cellcolor{gray!20}75.72&\cellcolor{gray!20}0.76&\cellcolor{gray!20}76.12&\cellcolor{gray!20}0.13&\cellcolor{gray!20}74.75&\cellcolor{gray!20}1.95
                         \\ \bottomrule[1pt]
\end{tabular}}
\label{tab:ra_res18_cifar100_5}
\end{table}

\begin{table}[t]
\centering
\caption{\small The post-purification robustness performance against diverse defense methods. We evaluate the performance on Tiny-ImageNet with the pretrained ResNet-18 and set the poisoning rate to $5\%$. The \textit{O-Backdoor} indicates the original performance of backdoor attacks, \textit{O-Robustness} metric represents the purification performance of the defense method, and the \textit{P-Robustness} metric denotes the post robustness after applying RA. All metrics are measured in percentage (\%).
}
\resizebox{0.8\textwidth}{!}{\begin{tabular}{c|c|cc|cc|cc}

\toprule[1pt]
\multirow{2}{*}{Method}  & \multirow{2}{*}{Mode} & \multicolumn{2}{c|}{BadNet} & \multicolumn{2}{c|}{Blended} & \multicolumn{2}{c}{SSBA}   \\ \cmidrule{3-8} 
                         &                                 & C-Acc($\uparrow$)            & ASR($\downarrow$)            & C-Acc($\uparrow$)            & ASR($\downarrow$)            & C-Acc($\uparrow$)             & ASR($\downarrow$)            \\ \midrule
Attack  & O-Backdoor &72.60&99.01&73.68&99.99&73.02&97.05\\
\midrule
\multirow{2}{*}{Clean}  & O-Robustness &73.88&0.08&73.88&0.10&73.88&0.05\\
                         & P-Robustness &72.99&0.76&73.48&0.60 &72.10&2.72
                         \\ \midrule
\multirow{2}{*}{EP} & O-Robustness    &70.90&0.02&71.11&0.01&70.24&0.01\\
                         & P-Robustness &70.59&0.65&70.93&0.09&69.98&1.90  \\ \midrule\midrule
\multirow{2}{*}{SAM}    & O-Robustness &70.65&5.92&70.80&4.02&70.18&11.22 \\
                         & P-Robustness  &70.90&84.64&71.30&99.41&70.61&77.85
                         \\ \midrule

\multirow{2}{*}{FST}    & O-Robustness  &65.36&0.63&66.55&1.19&65.85&3.32\\
                         & P-Robustness   &65.30&93.49&65.58&60.12&64.77&77.58
                         \\ \midrule

\multirow{2}{*}{BTI}    & O-Robustness &68.43&0.06&68.74&0.55&68.92&1.59\\
                         & P-Robustness  &68.91&68.29&68.41&38.13&67.98&77.25
\\ \midrule
\multirow{2}{*}{PAM (Ours)}    & O-Robustness &68.78&0.06&67.89&0.14&68.01&4.47\\
                         & P-Robustness  &\cellcolor{gray!20}68.97&\cellcolor{gray!20}7.81&\cellcolor{gray!20}66.86&\cellcolor{gray!20}0.37&\cellcolor{gray!20}67.39&\cellcolor{gray!20}14.26
                         \\ \bottomrule[1pt]
\end{tabular}}
\label{tab:ra_res18_tiny_5}
\end{table}

\begin{figure}
\centerline{\includegraphics[width=1\textwidth]{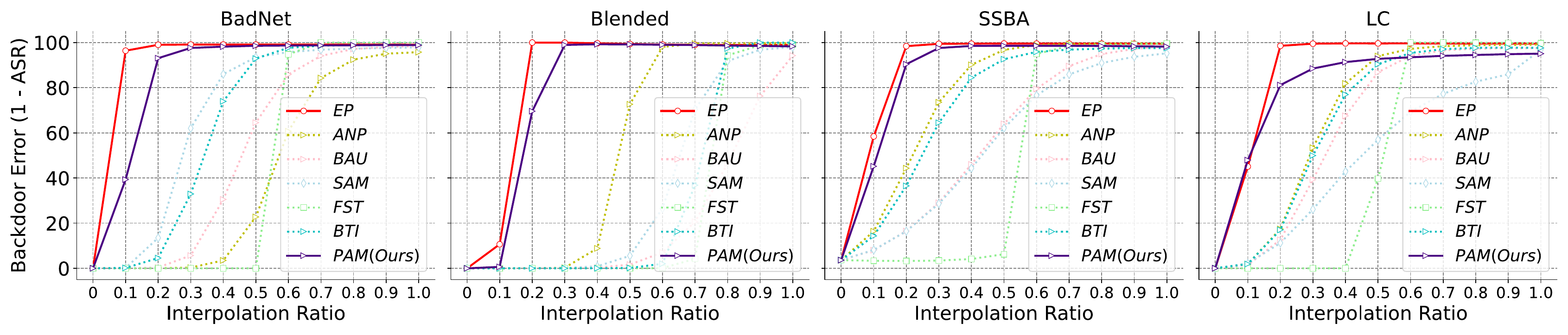}}
    \caption{The experimental results of LMC on CIFAR-10. We evaluate the performance on ResNet-18 and set the poisoning rate to $5\%$.}
    \label{fig:lmc_res18_cifar10_5}
\end{figure}

\begin{figure}
\centerline{\includegraphics[width=1\textwidth]{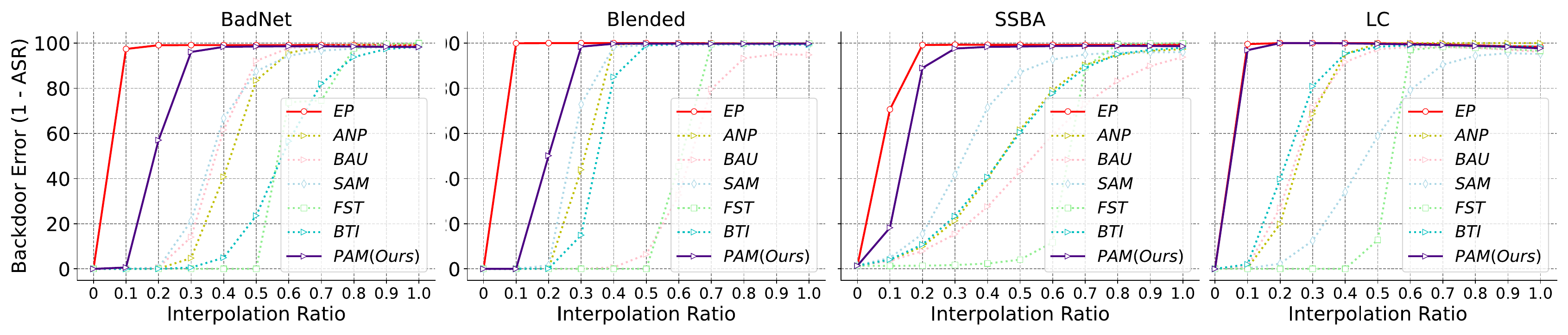}}
    \caption{The experimental results of LMC on CIFAR-10. We evaluate the performance on ResNet-18 and set the poisoning rate to $10\%$.}
    \label{fig:lmc_res18_cifar10_10}
\end{figure}

\begin{figure}
\centerline{\includegraphics[width=1\textwidth]{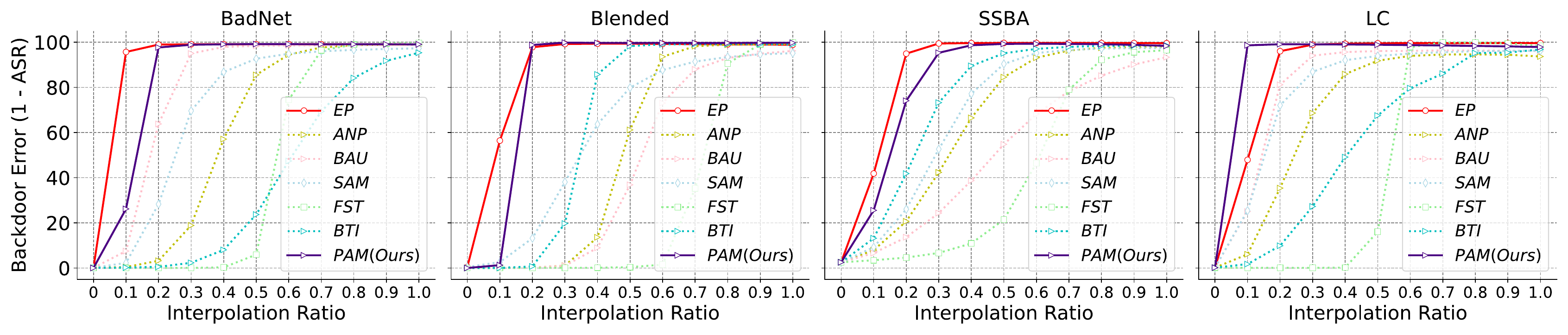}}
    \caption{The experimental results of LMC on CIFAR-10. We evaluate the performance on ResNet-50 and set the poisoning rate to $5\%$.}
    \label{fig:lmc_res50_cifar10_5}
\end{figure}

\begin{figure}
\centerline{\includegraphics[width=1\textwidth]{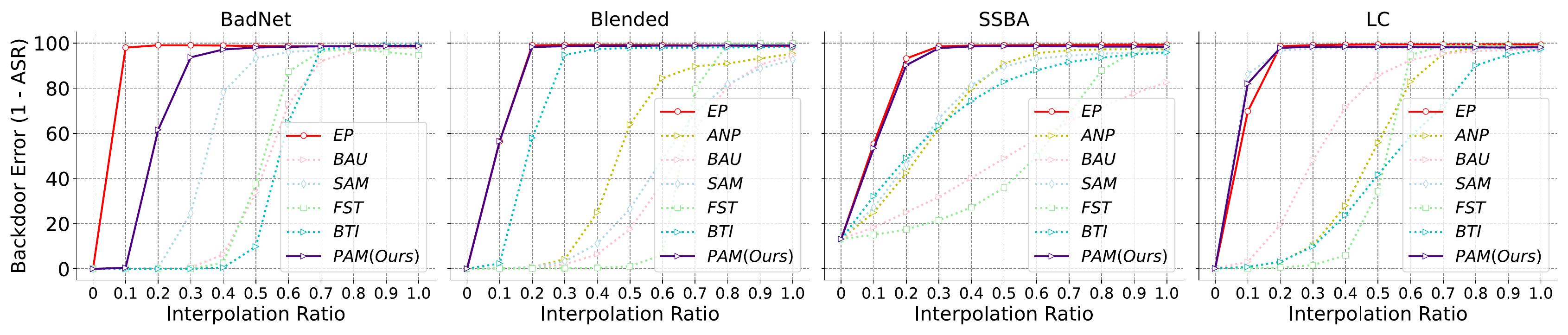}}
    \caption{The experimental results of LMC on CIFAR-10. We evaluate the performance on DenseNet-161 and set the poisoning rate to $5\%$.}
    \label{fig:lmc_dense_cifar10_5}
\end{figure}

\subsection{Detailed Results of LMC}
\label{app:detailed_lmc}
\textbf{Detailed LMC performance} In addition to the average LMC on CIFAR-10, we provide detailed experimental results for each attack setting. We demonstrate the LMC against ResNet-18 with $5\%$ poisoning rate in Figure~\ref{fig:lmc_res18_cifar10_5}; ResNet-18 with $10\%$ poisoning rate in Figure~\ref{fig:lmc_res18_cifar10_10}; ResNet-50 with $5\%$ poisoning rate in Figure~\ref{fig:lmc_res50_cifar10_5} and DenseNet-161 with $5\%$ poisoning rate in Figure~\ref{fig:lmc_dense_cifar10_5}, respectively. Our experiments demonstrate that the proposed PAM method effectively introduces high error barriers along the backdoor-connected path, leading to a greater deviation from the backdoored models.

\begin{figure}
\centerline{\includegraphics[width=1\textwidth]{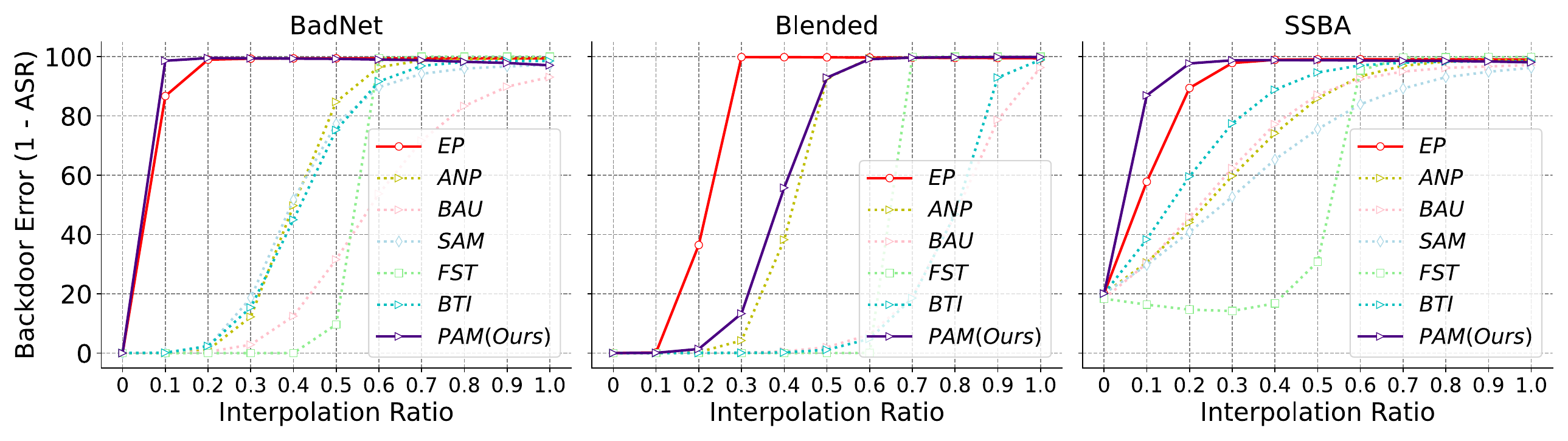}}
    \caption{The experimental results of LMC on CIFAR-10. We evaluate the performance on ResNet-18 and set the poisoning rate to $1\%$.}
    \label{fig:lmc_res18_cifar10_1}
\end{figure}

\textbf{LMC under Lower Poisoning Rate}
In this section, we evaluate the LMC under a lower poisoning rate, namely $1\%$. As shown in Figure~\ref{fig:lmc_res18_cifar10_1}, our PAM still achieves stable high error barriers along the backdoor-connected path compared to other defense methods.

\begin{figure}
\centerline{\includegraphics[width=1\textwidth]{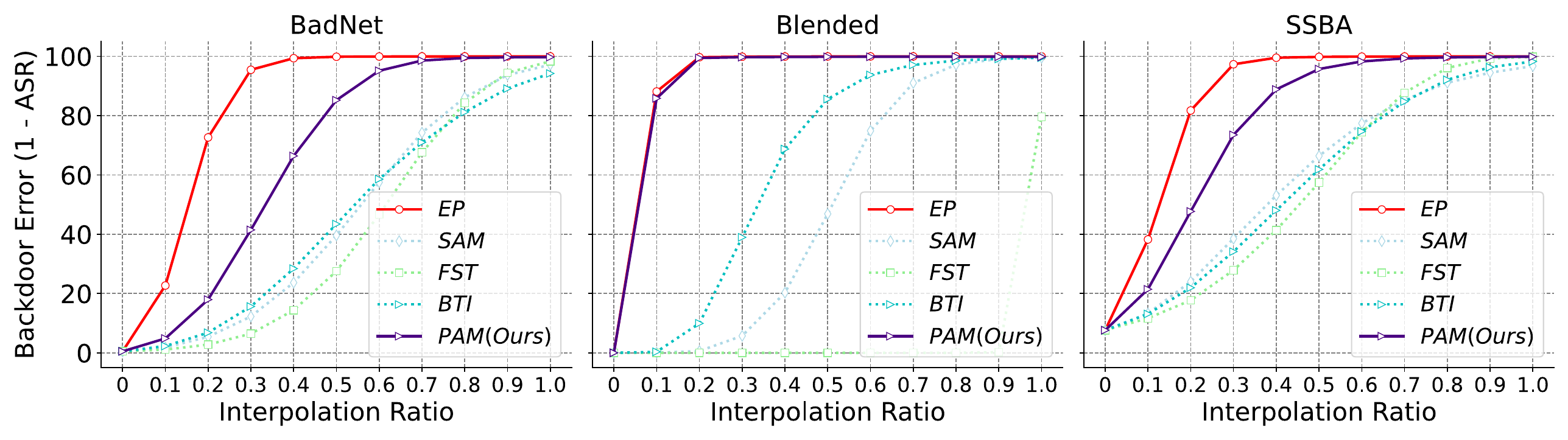}}
    \caption{The experimental results of LMC on CIFAR-100. We evaluate the performance on ResNet-18 and set the poisoning rate to $5\%$.}
    \label{fig:lmc_res18_cifar100_5}
\end{figure}

\begin{figure}
\centerline{\includegraphics[width=1\textwidth]{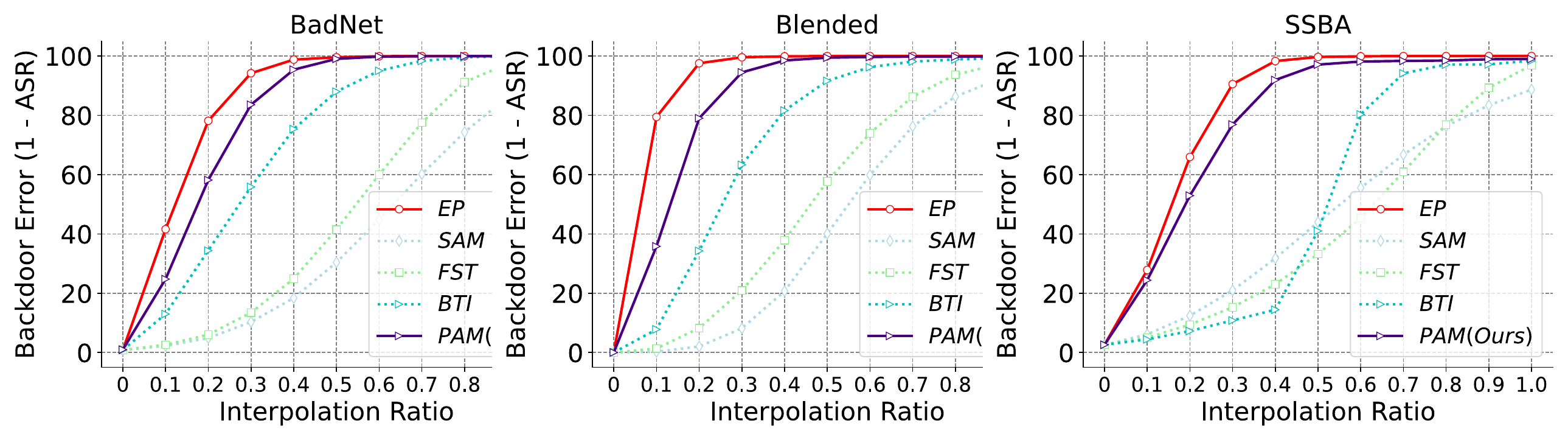}}
    \caption{The experimental results of LMC on Tiny-ImageNet. We evaluate the performance on ResNet-18 and set the poisoning rate to $5\%$.}
    \label{fig:lmc_res18_tiny_5}
\end{figure}

\textbf{LMC under Other Datasets}
We further evaluate the LMC on other two datasets, including the CIFAR-100 (shown in Figure~\ref{fig:lmc_res18_cifar100_5}) and the Tiny-ImageNet (shown in Figure~\ref{fig:lmc_res18_tiny_5}). Observations through the experimental results suggest that our PAM is also effective in introducing high error barriers along the backdoor-connected path across diverse datasets.

\begin{figure}
\centerline{\includegraphics[width=\textwidth]{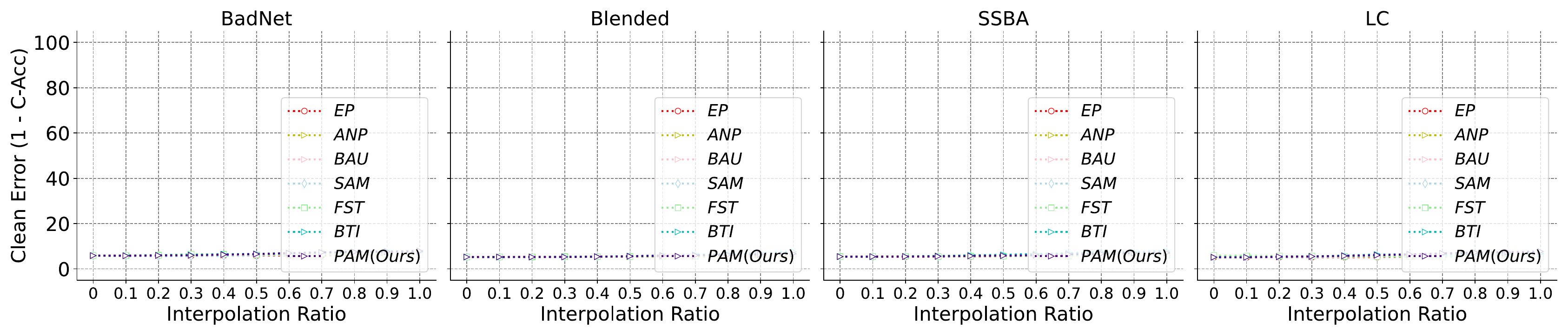}}
    \caption{The experimental results of the clean-connected path on CIFAR-10. We evaluate the performance on ResNet-18 and set the poisoning rate to $5\%$.}
    \label{fig:lmc_connected_clean}
\end{figure}

\textbf{LMC with Clean Samples}
In addition to evaluating the LMC with backdoored examples, we also provide experimental results of the LMC with clean samples, which we refer to as the clean-connected path. As depicted in Figure~\ref{fig:lmc_connected_clean}, we observe that diverse purification techniques exhibit almost no clean error barrier along the clean-connected path.

\begin{figure}
\centerline{\includegraphics[width=1\textwidth]{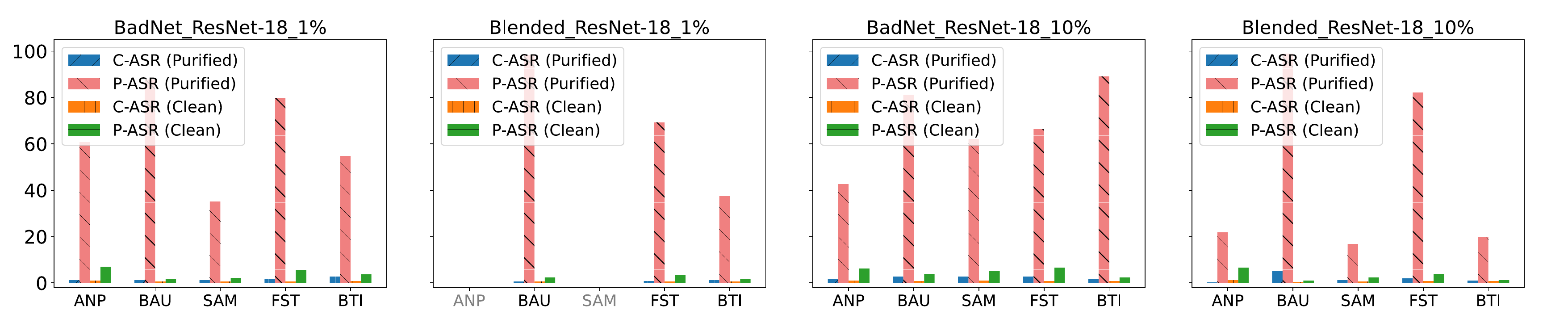}}
    \caption{Experimental results of QRA against two poisoning rates ($5\%$ and $10\%$).}
    \label{fig:ra_pratio}
\end{figure}

\begin{figure}
\centerline{\includegraphics[width=1\textwidth]{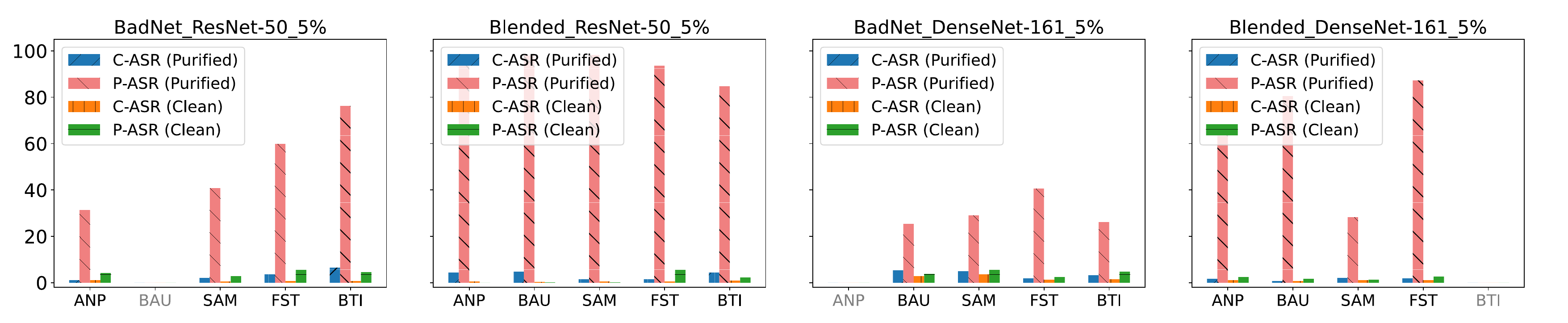}}
    \caption{Experimental results of QRA against two model architectures: ResNet-50 and DenseNet-161.}
    \label{fig:ra_archi}
\end{figure}

\subsection{Additional Results of QRA} 
\label{app:detailed_qra}
In this section, we evaluate the QRA under various attack configurations on the CIFAR-10 dataset. Specifically, we include additional poisoning rates ($1\%$ and $10\%$) in Figure~\ref{fig:ra_pratio}, and two model architectures (ResNet-50 and DenseNet-161) in Figure~\ref{fig:ra_archi}. We report the performance against two attacks, namely the BadNet and Blended. Note that for certain defense methods, their ASR after applying RA is low, making it hard to perform QRA evaluations on them. Therefore, we exclude these defense methods from our evaluation and represent them as light gray in the Figure. Experimental results demonstrate the effectiveness of our QRA on waking backdoor behavior under various attack settings.

\subsection{Results of Analyzing Sensitivity of $\rho$}
\label{app:results_rho}

We present the model performance across various $\rho$ values in Table \ref{tab:result_rho}. Our observations reveal that as $\rho$ rises, there is a slight decrease in clean accuracy alongside a significant enhancement in robustness against the RA. Additionally, we note that performance is relatively insensitive to $\rho$ when it exceeds 0.3. Given that we primarily monitor C-Acc (with the validation set) in practice, we aim to achieve a favorable trade-off between these two metrics. Therefore, we follow the approach of FST \cite{min2024towards} and select $\rho$ to ensure that C-Acc remains above a predefined threshold, such as $92\%$. 
\begin{table}[t]
\centering
\caption{\small We demonstrate the performance of PAM with diverse $\rho$ and evaluate the Blended attack on CIFAR-10 with ResNet-18. The O-Robustness metric represents the purification performance of the defense method, and the P-Robustness metric denotes the post robustness after applying RA.
}
\vspace{0.1cm}
\resizebox{\textwidth}{!}{\begin{tabular}{c|c|c|c|c|c}
\toprule[1pt]
Evaluation Mode & $\rho=0.1$ (C-Acc / ASR) & $\rho=0.3$ (C-Acc / ASR) & $\rho=0.5$ (C-Acc / ASR) & $\rho=0.7$ (C-Acc / ASR)& $\rho=0.9$ (C-Acc / ASR)                                    \\ \midrule
O-Robustness    & 94.03 / 6.33  & 93.64 / 2.07  & 93.34 / 1.67 &  92.12 / 0.50 & 91.99 / 1.00 \\ \midrule
P-Robustness    & 93.60 / 33.29 & 93.61 / 10.06 & 93.38 / 2.69 & 92.17 / 2.62 & 92.54 / 0.30 \\
\bottomrule[1pt]
\end{tabular}}
\label{tab:result_rho}
\end{table}

\subsection{Additional Discussions on the Frequently Asked Question: Directly Applying SAM with BTI} 
\label{app:sam_pam}
Given that our proposed method exhibits some similarities to the functionality of SAM, a frequently asked question arises regarding whether utilizing reversed backdoor data from BTI in conjunction with SAM would enhance post-purification robustness. To address this question, we further evaluate the performance of directly integrating SAM with BTI (BTI+SAM) as a robust baseline and present a comparative analysis with PAM in Table~\ref{tab:sam_pam}. Our experimental results indicate that the direct combination of BTI and SAM does not achieve the same level of robustness as PAM. This finding underscores the significance of the backdoor-connected pathway between purified and backdoored models, as delineated by the LMC.

\begin{table}[h]
\centering
\caption{\small The comparison of post-purification robustness performance between BTI+SAM and PAM. We evaluate the performance on CIFAR-10 with ResNet-18 and set the overall poisoning rate to $5\%$. The \textit{O-Backdoor} indicates the original performance of backdoor attacks, \textit{O-Robustness} metric represents the purification performance of the defense method, and the \textit{P-Robustness} metric denotes the post robustness after applying RA. All metrics are measured in percentage (\%).
}
\resizebox{\textwidth}{!}{\begin{tabular}{c|c|cc|cc|cc|cc}

\toprule[1pt]
\multirow{2}{*}{Method}  & \multirow{2}{*}{Mode} & \multicolumn{2}{c|}{BadNet} & \multicolumn{2}{c|}{Blended} & \multicolumn{2}{c|}{SSBA} & \multicolumn{2}{c}{LC}  \\ \cmidrule{3-10} 
                         &                                 & C-Acc($\uparrow$)            & ASR($\downarrow$)            & C-Acc($\uparrow$)            & ASR($\downarrow$)            & C-Acc($\uparrow$)             & ASR($\downarrow$)            & C-Acc($\uparrow$)          & ASR($\downarrow$)      \\ \midrule
Attack  & O-Backdoor &94.04&99.99&94.77&100.0&94.60&96.38&94.86&99.99\\
\midrule

\multirow{2}{*}{BTI+SAM}    & O-Robustness &91.72&3.70&92.49&42.49&93.03&7.84&93.04&6.67\\
                         & P-Robustness  &92.48&100.0&92.99&100.0&93.36&94.12&92.87&92.01

\\ \midrule
\multirow{2}{*}{PAM (Ours)}    & O-Robustness &92.11&1.14&93.34&1.67&92.96&1.24&92.32&4.92
\\
& P-Robustness &\cellcolor{gray!20}91.66&\cellcolor{gray!20}3.90&\cellcolor{gray!20}93.38&\cellcolor{gray!20}2.69&\cellcolor{gray!20}92.20&\cellcolor{gray!20}3.31&\cellcolor{gray!20}92.15&\cellcolor{gray!20}8.31
                         \\ \bottomrule[1pt]

\end{tabular}}
\label{tab:sam_pam}
\end{table}

\clearpage
\newpage

\end{document}